\documentclass{article} 
\usepackage[OT1]{fontenc} 
\usepackage{iclr2026_conference,times}

\usepackage{hyperref}
\usepackage{url}
\usepackage{booktabs}
\usepackage{multirow}
\usepackage{amsmath,amssymb}
\usepackage{graphicx}
\usepackage{xcolor}
\usepackage{enumitem}
\usepackage{caption}
\hypersetup{hidelinks,
  pdftitle={Compute Globally, Materialize Locally: The Memory Contract of Sparse Event-KV},
  pdfauthor={Zefeng Cai, Zerui Cai}}

\newcommand{\pp}{\,\mathrm{pp}}
\newcommand{\arm}[1]{\texttt{#1}}
\newcommand{\kv}{\mathrm{KV}}
\newcommand{\serve}{\Sigma}
\newcommand{\esrc}{e_{\mathrm{src}}}
\newcommand{\eroot}{e_{\mathrm{root}}}
\newcommand{\eedge}{e_{\mathrm{edge}}}

\title{Compute Globally, Materialize Locally:\\
\large The Memory Contract of Sparse Event-KV}

\author{Zefeng Cai, \texttt{caizefeng994@gmail.com}\\
Zerui Cai, \texttt{zrcai\_flow@126.com}\\
Independent Researchers\\[0.3em]
\url{https://github.com/oklen/Compute-Globally-Materialize-Locally}}

\iclrfinalcopy

\begin{document}
\maketitle
\lhead{Preprint --- July 2026; revised October 2026}

\begin{abstract}
Long-horizon agents increasingly reuse their KV cache as memory, keeping some entries and dropping the rest on the rarely tested premise that a retained event stays informative once the observations that produced it are gone. We omit one earlier observation from what is served, across otherwise identical agent histories. Among items sensitive to that observation, the answer overwhelmingly follows the omitted value, though no served span says which value is correct: a downstream event's cached rows act as an independently servable view of computation whose inputs are gone, which we call semantic materialization. It can also be written on purpose: an answer-free event raises donor-aligned recovery from 6\% to 51\% on Qwen3-8B, whereas harvesting natural mentions from long-term dialog yields no detected advantage. What the model reads back is bounded: compact state comes back while larger payloads decay toward chance, and phrasing, not meaning alone, decides whether a state is read back at all, though a linear probe finds the state in the carrier row under every phrasing. Such rows are also hard to address: deployed retrieval cannot tell the event that carries a value from the one that points at it, and on real dialogs it finds the answer-bearing turn for single-hop questions but only at chance for multi-hop ones. The result is a memory contract for sparse event-KV serving: what to write, where it lands, and what survives once the source is gone. Dropping a source without losing accuracy thus does not show it was unnecessary, and two deployed heuristics invert: attention at eviction marks what the agent has already written down and so no longer needs, and citations fail mostly by omission, not decoration.
\end{abstract}

\section{Introduction}
\label{sec:intro}

Long-horizon agents increasingly keep their raw interaction history --- and, more
aggressively, the KV cache produced while reading it --- as their memory substrate.
Serving that memory is necessarily \emph{sparse}: a future query touches a handful of
events, so systems select a few spans (rows of the cache) and drop the rest
\citep{h2o2023,snapkv2024,epicache2025} --- including, routinely, the observations that
generated the values the query asks about. Eviction and episodic-memory policies are validated on
downstream accuracy, but those evaluations do not isolate whether a \emph{retained} span still
carries state derived from an \emph{evicted} one. KV memory makes a bet that text retrieval cannot:
during prefill later tokens are \emph{contextualized} with access to the source, so their cache rows
may already contain the result of computation whose inputs will never be served again.

\begin{figure}[t]
\centering
\includegraphics[width=\linewidth]{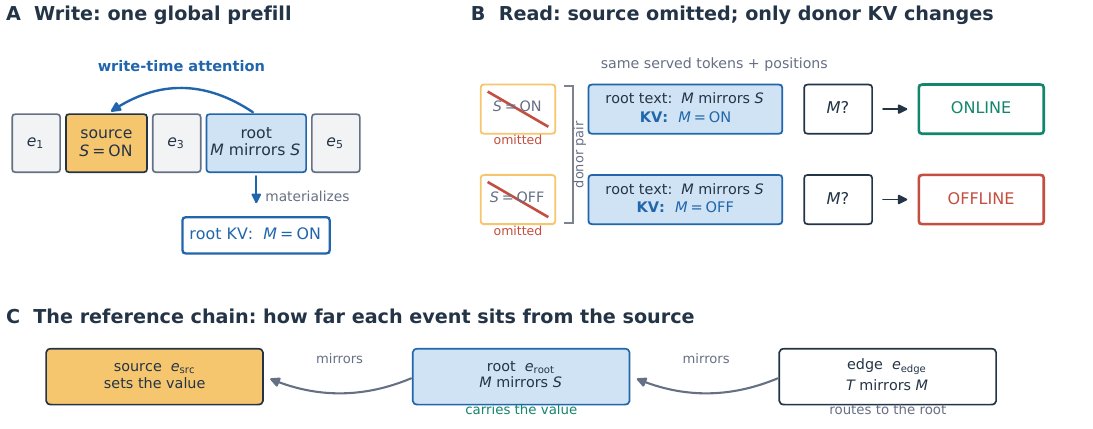}
\caption{\textbf{Semantic materialization, and the comparison that isolates it.}
\textbf{A}: during one global prefill, a downstream \emph{root} event (``$M$ mirrors $S$'',
which names no value) attends to the \emph{source} that set $S$, so its cache rows come to
encode $S$'s state --- the root becomes a donor-dependent representation of $S$. \textbf{B}: at
serve time we omit the source and serve the root --- alongside a fixed donor-invariant decoy --- and the query. A \emph{donor pair} holds
every served token and position fixed and flips only the omitted source; the answer follows that
omitted value, so the served rows carry more than their visible text. \textbf{C}: the reference chain
--- the \emph{root} mirrors the \emph{source} and carries its value, while a \emph{reference
edge} mirrors the root, one hop further from the source, and mostly routes a query back to it.}
\label{fig:concept}
\end{figure}

This paper measures that bet with one controlled comparison (Fig.~\ref{fig:concept}). One global
prefill gives each event $e_i$ of a history $\mathcal{T}=(e_1,\dots,e_n)$ cache rows $\kv(e_i)$
contextualized by the events before it. Serving keeps a subset $\serve\subseteq\mathcal{T}$ and drops
the rest --- including the \emph{source} event $\esrc$ that wrote the value $v$ we then ask about.
What survives is a downstream \emph{root} event $\eroot$ (``$M$ mirrors $S$'') whose own text names no
value; we serve it, drop $\esrc$, and ask for $M$'s state. To make the test causal we compare a
\emph{donor pair}: two histories byte-identical in every served token and position, differing only
in whether $\esrc$ wrote $v$ or $v'$. The answer follows the omitted value \textbf{99:0} on Qwen3-8B,
though the served text never states it (\S\ref{sec:discovery}): selecting which cache rows to keep
--- the operation every eviction or memory policy performs --- does not select only their visible
token content.

Our central finding, and its boundary:

\begin{quote}
\emph{Global prefill can causally materialize compact, source-omitted state into
selected downstream event spans. Deliberately written events expose this primitive under
calibrated constructions and readouts, whereas passive natural harvesting is not a
dependable end-to-end interface.}
\end{quote}

We call the phenomenon \textbf{semantic materialization}: \emph{the write-time commitment of an
upstream-derived state into a downstream event representation, without requiring the event tokens to
repeat that state verbatim.} Prior work shows that prefill can write memoized conclusions into
downstream notes \citep{mtn2026}; the open serving question is what an \emph{independently served}
span carries once its \emph{source is dropped} and the exact future query was absent at write time.
Accuracy alone cannot answer it --- the served text may contain the answer, the model may guess from
priors or eliminate decoys --- so every comparison here is a donor pair, with decoys sampled
independently (App.~\ref{app:ext-intro}). Our contributions are the phenomenon and the contract that
governs it:

\begin{enumerate}[leftmargin=1.4em,itemsep=1pt,topsep=2pt]
\item \emph{Selecting rows does not select only their visible token content.} Among donor-sensitive
items the answer follows the omitted source \textbf{99:0} on Qwen3-8B, though no served span states
which value is correct, and the direction holds across serving cells, models and readouts
(\S\ref{sec:discovery}).
\item \emph{The contract that governs it: trigger, landing, access.} Surface form rather than meaning
alone decides whether a carried state is read back: over a 16-construction bank comprehended at
$.98$, write-through runs from chance to $.95$, yet a linear probe recovers the bound state from the
carrier row at $.99$--$1.00$ under every construction on four checkpoints (\S\ref{sec:write}). The
root carries the state while served reference edges mostly route a query back to it
(\S\ref{sec:compose}), and the readout reaches a compact-state envelope while larger payloads stay
linearly present in the served rows (\S\ref{sec:envelope}).
\item \emph{Capacity is not addressability.} An index over event text cannot separate the row that
carries a value from the row that points at it, and on real long-term dialogs a full deployed stack
finds the answer-bearing turn well on single-hop questions and no better than chance on multi-hop
ones (\S\ref{sec:address}).
\item \emph{The primitive is programmable; passive harvesting is not.} A deliberately written,
answer-free carrier raises donor-aligned recovery from $6\%$ to $51\%$ on Qwen3-8B, whereas natural
mentions --- including the rows a deployed retrieval stack selects when the answer-bearing turn is
withheld ($n{=}1269$, $|\delta|$ within about $\pm.03$) --- carry nothing of the withheld value under
a same-path positive control that fires. What a harvested row does carry is its own context, with
$1.9$ log-odds more confidence than a re-encoding of the same text at ceiling accuracy: fidelity, not
accuracy (\S\ref{sec:active}).
\item \emph{Two deployed heuristics, measured against the contract.} On agent trajectories the
attention available before the next query marks only what the agent already wrote down, and citation
checking misses the dominant failure, omitted load-bearing passages ($20.7\%$, against $1.5\%$
decorative). An ablation that drops a source and loses no accuracy has not shown the source was
unnecessary; on these trajectories the value usually survives as a textual copy
(\S\ref{sec:consequences}).
\end{enumerate}

\section{The serving contract}
\label{sec:setup}

\textbf{Three rules make the question non-trivial.} The donor comparison of \S\ref{sec:intro}
obeys a \emph{serving contract}: (i) the \emph{source} events that defined the queried state are
excluded from the served set $\serve$; (ii) the exact future query instance and its wording are
absent at write time, though a predicate/target schema may be anticipated; (iii) we allow no
auxiliary trained reader and no fine-tuning, so the frozen model must recover state through its
native interface. Span selection uses prespecified event roles in the controlled trajectories
and lexical heuristics on real dialogs.

\textbf{Three predicates, used throughout.} A served span $e$ \emph{materializes} $v$ for a readout
$q$ when its contextualized rows $\kv(e)$ induce donor-aligned behavior relative to isolated
re-encoding of the same text, with $\esrc$ omitted and $e$'s visible tokens and positions held fixed.
The \emph{trigger} is the write-time construction that makes an event do this, and a \emph{carrier}
is such an event --- one whose $\kv$ holds upstream-derived state without naming it in text. A served
span is a \emph{conditional landing} when it materializes $v$ while selected alongside fixed
donor-invariant scaffolding (a decoy, and an unrelated register in the active arms); this is what
separates roots, which land the value, from reference edges, which mostly route to it.

\textbf{Where the query goes.} Retained rows always keep their original positions; only the query's
placement is a protocol choice. Our synthetic arms append the query after the last retained row,
whereas an eviction-style system leaves it on the original timeline \citep{snapkv2024}. Placement
changes effect size --- a donor-invariant filler at the mention slot costs $-.198$ on Qwen3
(App.~\ref{app:poscontrol}) --- and sliding-window checkpoints are served at original positions
(App.~\ref{app:swgate}; App.~\ref{app:ext-setup} states the choice in full). Protocols, the operative
null and statistical conventions are in Appendices~\ref{app:protocols} and~\ref{app:honest}.

\section{Discovery: a served row answers from an omitted source}
\label{sec:discovery}

\begin{table}[t]
\centering
\caption{\textbf{Donor-sensitive answers overwhelmingly follow the hidden donor.}
Follow:anti counts among scorable donor-dependent pairs (differing binary outputs), source omitted, across the serving cells and
models defined in the text. Exact two-sided sign tests; Gemma-4 read out under the menu protocol.}
\label{tab:donor}
\begin{tabular}{llc}
\toprule
Checkpoint & Serving cell & follow\,:\,anti \\
\midrule
Qwen3-8B (2025)          & root only          & \textbf{99:0} \\
Qwen3-8B                 & mention absent      & 52:1 \\
Ministral-3-8B (2025)    & mention absent      & 90:1 \\
Gemma-4-12B (2026)       & mention co-served   & 80:0 \\
\bottomrule
\end{tabular}
\end{table}

Concretely (Fig.~\ref{fig:concept}): one event writes ``register $S$ is \textsc{online}'', and a
later event states only ``$M$ mirrors $S$''. We omit the first, serve the second, and ask for $M$ ---
nothing served names \textsc{online}, yet the model answers \textsc{online}, and switches to
\textsc{offline} when we flip the omitted write. Across many such donor pairs we count, among
\emph{donor-sensitive} items, how many \emph{follow} the omitted value versus its opposite
(\emph{anti}; App.~\ref{app:protocols}). On Qwen3-8B the split is \textbf{99:0}
($p{=}3.2{\times}10^{-30}$); a $256$-pair replication finds $131$ donor-dependent pairs and a
direction of $130{:}0$ ($p{=}1.5{\times}10^{-39}$). The direction holds across serving cells, models
and readouts: no cell of Table~\ref{tab:donor} shows more than one \textsf{anti} pair, the narrowest
root cell measured anywhere is still $39{:}18$ (Fig.~\ref{fig:compose}A), and under the
fixed-candidate menu that Gemma-4 requires Qwen3 still follows the omitted source $88{:}0$
(App.~\ref{app:ext-discovery}).

\emph{Selecting rows does not select only their visible token content:
contextualized event rows can carry materialized results of omitted computation.}

\textbf{What crosses is the state, not necessarily the string.} Semantic polarity transfers on every
checkpoint tested (follow:anti $56{:}0$, $45{:}0$, $38{:}0$ on Qwen3, Ministral-3, Gemma-4), while the
donor's exact word transfers on Gemma-4 ($52{:}0$) but not on Qwen3 ($0{:}0$; verbatim $9.7\%$ against
an $8\%$ control) or Ministral-3 ($4{:}0$; App.~\ref{app:wording}). An explicit later update overrides
what was carried over: no old-value leakage ($0/96$), and the served update wins every time ($96/96$;
App.~\ref{app:ext-discovery}).

\section{Trigger: which events materialize the state}
\label{sec:write}

Does every event that \emph{reads} an upstream state also \emph{write} it in a form the model reads back? No. We sweep sixteen
carrier constructions in two families (Table~\ref{tab:bank}): eight \emph{mirror} constructions
``register $M$ $\langle\text{relation}\rangle$ register $S$'' and eight \emph{flag} constructions
``the alert flag for sensor $r$ $\langle\text{relation}\rangle$ the check result'', neither naming
a value. Each construction is scored on two \emph{rates} over its $64$ donor items: \emph{comprehension} (correct from
the full source text) and \emph{write-through} (correct from the carrier's rows once the source is
dropped). Comprehension sits at or near ceiling on every recent-model construction (mean $.98$ over the
$48$ cells, minimum $.77$) and does not predict write-through, so a construction's \emph{readout class} is set
by write-through --- \textbf{W} reads back ($\ge.75$), \textbf{U} does not ($<.625$), \textbf{P}
partial in between; \textbf{X} marks failure of the full-text comprehension control and is
excluded from write-through classification (no recent model is \textbf{X}). Write-through is an output-level rate: regularity (4) shows the carrier row holds the state under every construction, so these classes sort constructions by whether the model reads that state back.

\begin{table}[t]
\centering
\caption{\textbf{The sixteen-construction write bank}, in two families. \emph{Mirror}
constructions read ``register $M$ $\langle$relation$\rangle$ register $S$''; \emph{flag}
constructions read ``the alert flag for sensor $r$ $\langle$relation$\rangle$ the check result''
--- neither naming a value. Write-through rate and class per model (serve carrier rows only,
source omitted; $64$ donor items each; comprehension mean $.98$, minimum $.77$). No construction is class
\textbf{W} on all three models, and no Gemma-4 construction reaches \textbf{W} under this
trace-generation readout; totals (W/P/U) are $5/4/7$ Qwen3, $7/2/7$
Ministral-3, $0/2/14$ Gemma-4.}
\label{tab:bank}
\small
\setlength{\tabcolsep}{6pt}
\begin{tabular}{llccc}
\toprule
constr. & relation & Qwen3 & Ministral-3 & Gemma-4 \\
\midrule
\multicolumn{5}{l}{\emph{Mirror family} --- ``register $M$ $\langle$relation$\rangle$ register $S$''} \\
\texttt{follows}  & follows                  & .83\,W & .81\,W & .55\,U \\
\texttt{mirrors}  & mirrors                  & .80\,W & .89\,W & .52\,U \\
\texttt{copies}   & copies                   & .78\,W & .77\,W & .55\,U \\
\texttt{shadows}  & shadows                  & .78\,W & .95\,W & .50\,U \\
\texttt{kept\_eq} & is kept equal to         & .72\,P & .84\,W & .58\,U \\
\texttt{matches}  & matches                  & .64\,P & .94\,W & .67\,P \\
\texttt{tracks}   & tracks                   & .55\,U & .45\,U & .66\,P \\
\texttt{synced}   & is synchronized with     & .45\,U & .48\,U & .56\,U \\
\addlinespace
\multicolumn{5}{l}{\emph{Flag family} --- ``the alert flag for sensor $r$ $\langle$relation$\rangle$ the check result''} \\
\texttt{consist}  & was made consistent with & .94\,W & .95\,W & .47\,U \\
\texttt{inline}   & was updated in line with & .69\,P & .56\,U & .42\,U \\
\texttt{accord}   & was set according to     & .64\,P & .62\,P & .45\,U \\
\texttt{record}   & now records              & .58\,U & .53\,U & .45\,U \\
\texttt{reflect}  & was set to reflect       & .52\,U & .64\,P & .53\,U \\
\texttt{update}   & was updated accordingly  & .52\,U & .52\,U & .44\,U \\
\texttt{write}    & was written from         & .52\,U & .56\,U & .41\,U \\
\texttt{assign}   & was assigned by threshold & .48\,U & .53\,U & .47\,U \\
\bottomrule
\end{tabular}
\end{table}

Four regularities.

\textbf{(1) Same semantics, different fate.} Semantically equivalent constructions that the model
comprehends perfectly differ by tens of points in write-through: on Qwen3 \texttt{consist} reads back
at $.94$ but \texttt{write} at only $.52$, and Ministral-3 splits the same pair $.95$ versus $.56$. The
wording itself carries this: the profile holds when every construction shares one background and query
position ($\rho{=}.88$) and across two checkpoints with independent backgrounds ($\rho{=}.78$;
App.~\ref{app:ext-write}).

\textbf{(2) No universal construction; the readout aperture differs by model.} The set of
constructions whose state the model reads back swings by checkpoint --- W/P/U totals $5/4/7$ on
Qwen3, $7/2/7$ on Ministral-3, $0/2/14$ on Gemma-4 --- so no construction is class W on all three.

\textbf{(3) Generation under-reports the logit.} A forced-choice output-logit probe finds the
candidate logit reachable where free generation does not realize it: on Gemma-4 for $8$ of the $16$
constructions, none of them class W. The free-generation interface is a \emph{model $\times$
construction $\times$ protocol} property.

\textbf{(4) The carrier row holds the state under every construction; the readout is what varies.} A representation-level test (App.~\ref{app:repprobe}) separates writing from reading. Two sources hold opposite states and the carrier names one of them at random, so its target is an exclusive-or that a linear probe can read only off a row that computed the named source's state. On all sixteen constructions and four checkpoints (the three above plus Qwen3-32B), a probe on the carrier row's served values recovers that state at $.99$--$1.00$; the row just before the carrier, which saw both sources but not which one is named, stays at its floor ($\le.63$); the answer position holds the state too ($.92$--$1.00$); and the model's own generated answer runs from $.39$ to $.98$. Linear decodability does not show that the model can use the state, but it places the construction profile at the readout: the wording decides whether the model says what the row holds, not whether the row holds it.

Together the sweep supports calibration rather than universality --- \emph{the readout of materialized
state must be calibrated per model} --- with an explicit textual record the one arm that recovers
on every model evaluated (\S\ref{sec:active}).

\section{Landing: roots carry, reference edges route}
\label{sec:compose}

\begin{figure}[t]
\centering
\includegraphics[width=\linewidth]{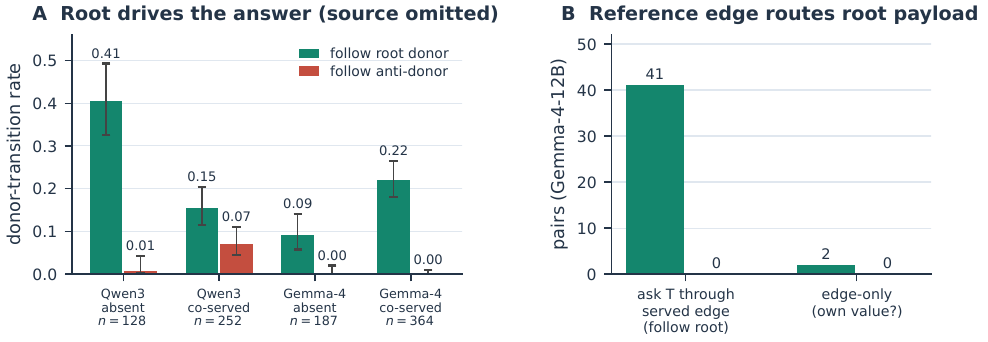}
\caption{\textbf{Roots carry the dominant donor-aligned signal; served reference
edges mostly route it (sparse serving, source omitted).}
\textbf{A}: the root-donor follow/anti \emph{rates} (Wilson $95\%$ CIs) in all four cells ---
mention-absent and mention-co-served, per checkpoint. Bars are rates over the stated denominators,
not counts: mention-absent cells contribute one root contrast per base item ($n{=}128$/$187$),
co-served factorial cells two ($n{=}252$/$364$), so heights are comparable only as rates. The
mention's own donor is inert on Qwen3 but has a substantial donor effect on Gemma-4. \textbf{B}:
with a true referencing edge (``$T$ mirrors $M$'') served, answers about $T$ follow the
\emph{root} donor 41:0 on Gemma-4-12B; the edge alone carries far less (Gemma-4 2:0; weaker
edge-local carryover on Ministral-3, 33:16). Panel A's four cells still differ in serve set and
appended query position, so cross-cell comparisons remain descriptive.}
\label{fig:compose}
\end{figure}

Where does the materialized value live when several later events attended the source? Roots carry
it; reference edges route to them. After the root ``$M$ mirrors $S$'', a downstream \emph{edge}
$\eedge$ (``$T$ mirrors $M$'') is asked about $T$ with the source omitted: Gemma-4 answers with the
\emph{root}'s donor $41{:}0$, yet the edge selected \emph{without} the root carries almost nothing
($2{:}0$). Ministral-3 reproduces the routing at $298{:}0$ with a weaker edge-local signal ($33{:}16$),
so ``edges route'' is the strong pattern rather than a law (Fig.~\ref{fig:compose}B). Served link by
link, a depth-3 chain on Qwen3 answers only from its root, and dropping the root collapses it from
$.707$ to $.465$ (App.~\ref{app:tables}).

A source-omitted $2{\times}2$ factorial that swaps the root's donor and a downstream \emph{mention}'s
donor independently finds the answer following the root on both checkpoints, sized differently by the
two serving protocols (Table~\ref{tab:factorial}). On Qwen3 that difference is chiefly position: a
donor-invariant filler at the mention slot accounts for $-.198$ of the $-.250$ confounded contrast
(App.~\ref{app:poscontrol}); on Gemma-4 the mention's own donor drives the answer. The root/edge
distinction is about role rather than position: compressing their slot gap from five slots to one
changes it by only $+.039$, and the contrast reproduces under free generation ($.398$ vs $.141$,
$p{=}1.3{\times}10^{-5}$; App.~\ref{app:ext-compose}).

\section{Access: binary state recovers, larger payloads fade}
\label{sec:envelope}

\begin{figure}[t]
\centering
\includegraphics[width=\linewidth]{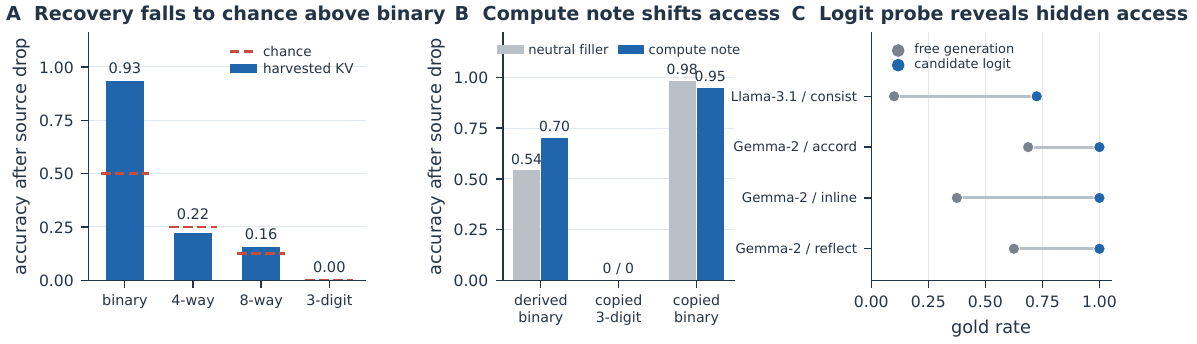}
\caption{\textbf{Native recovery is strong for binary state and falls toward chance as payload
cardinality grows; exact numeric recovery fails.} \textbf{A--B}: Qwen3-8B under one menu-form
readout with independent decoys. \textbf{A}: harvest-after-drop accuracy by payload type --- binary
well above chance, four- and eight-way toward it, three-digit at zero. \textbf{B}: a write-time
compute note shifts the copied$\to$derived frontier but not numeric payloads. \textbf{C}: the
generation--logit gap on selected legacy model--construction pairs (Llama-3.1, Gemma-2), where free
generation under-reports the candidate-logit readout.}
\label{fig:envelope}
\end{figure}

How much of this carried-over state does native recovery reach? Under a single menu-form readout
with independent decoys, a binary state recovers well above chance ($.934$, $95\%$ CI
$[.896,.958]$, against $.5$). Four-way lands at $.223$ ($90\%$ CI $[.183,.268]$) and eight-way at
$.156$ ($[.123,.197]$), against chance rates of $.25$ and $.125$; on a four-way payload two
orthogonal binary probes recover no more than the four-way menu does, so this is not an artifact of
four-way decoding (App.~\ref{app:tables}). Three-digit numbers are never
recovered exactly ($0/192$, $95\%$ upper $.02$).\footnote{Categorical cells
$n{=}256$, numeric/derived cells $n{=}192$. Each multi-way cell individually passes the prespecified $\pm.075$ TOST (the
eight-way borderline), but simultaneous equivalence across cardinalities is not established
under Holm correction, and on four-way the model echoes the served decoy.} \emph{Derived} binary
verdicts are not natively recovered above chance either: absent a write-time prompt we detect no
self-materialized conclusion at the readout interface (\S\ref{sec:limits}).

Recall rather than decoy-exclusion drives the binary result (about four-fifths of the successes;
App.~\ref{app:honest}); a write-time compute note moves derived verdicts ($.542\to.703$) but leaves
source-omitted numeric payloads at zero; and query form matters as much as the store --- several
carriers answer a source-addressed query but stay weak under a target-addressed one, so the primitive
is not reliably target-addressable (App.~\ref{app:ext-envelope}).

\textbf{The envelope bounds the readout, not the rows.} Linear probes on the served values of matched synthetic payload items (App.~\ref{app:repprobe}) recover binary, four- and eight-way payloads at $\approx\!1.00$ on all three recent models, the hundreds digit of a three-digit payload at $.45$--$.76$ (chance about $.1$), and whether the number exceeds $500$ at $.93$--$.97$ --- while the model's own answer to that last question sits at $.42$--$.59$. An unrelated row served from the same prefill decodes the payload as well, so the value is spread over the rows downstream of its source rather than confined to the carrier, and decodability is not use: what the envelope measures is how much of the carried value the model's readout reaches.

\section{Capacity is not addressability}
\label{sec:address}

\S\ref{sec:envelope} closes on an addressing problem internal to the query. A serving system faces
the external one: it is never handed the rows that carry the value, and must find them in an index
built over event \emph{text}. Everything above sets $P(\text{hit}){=}1$ by construction; here we
remove that assumption, first on the donor-paired register trajectories and then on real dialogs.

\textbf{What the model separates, a text index cannot.} On the index side, ``$A$ mirrors $B$'' and
``$B$ mirrors $A$'' are the same bag of words: restricted to the query's terms, a root and the edge
pointing at it carry \emph{identical} term multisets, so any scorer that is a function of the term
multiset --- Okapi BM25 included --- scores them bit-identically, the tie falls to recency, and the
edge wins on $200/200$ items. No wording makes a text index represent the direction of a reference
edge. On the model side the two are plainly different: with items, budget and scaffold fixed, serving
the root separates donors at $.398$ against $.141$ for the edge ($45{:}12$, exact McNemar
$p{=}1.3{\times}10^{-5}$), and the gap survives compressing the slot distance. A dense retriever finds
a one-hop carrier ($r@1{=}.83$) but ranks a multi-hop carrier below an irrelevant filler, and two
evaluation traps transfer off the template: a retriever that serves the source wins without
exercising materialization, and an index queried with the answer options measures a string match
(App.~\ref{app:ext-address}).

\textbf{The same question on real dialogs.} On LoCoMo \citep{locomo2024} --- $555$ questions over ten
conversations of $419$--$689$ turns --- we ask whether an index ranks a topical, answer-free turn above
every answer-stating turn, relative to chance. The stack agent-memory systems deploy (Okapi BM25,
Qwen3-Embedding-0.6B, reciprocal-rank fusion, and \texttt{bge-reranker-v2-m3} over the fused top-$50$)
beats chance overall, the full stack most ($-.42$), and does best on single-hop questions ($-.50$); on
multi-hop questions no arm is below chance (reranked fusion $-.008$ $[-.114,+.174]$) and BM25 actively
prefers the topical turn ($+.264$; Table~\ref{tab:e1}). A directed graph index failed its instrument
gate, so the claim is about text indexes, not about retrieval. The practical reading is a scope
statement: materialized rows pay where the source is genuinely gone, evicted or compacted; where it
merely has not been retrieved, fetching it is available and better.

\section{Programming the primitive}
\label{sec:active}

\begin{figure}[t]
\centering
\includegraphics[width=\linewidth]{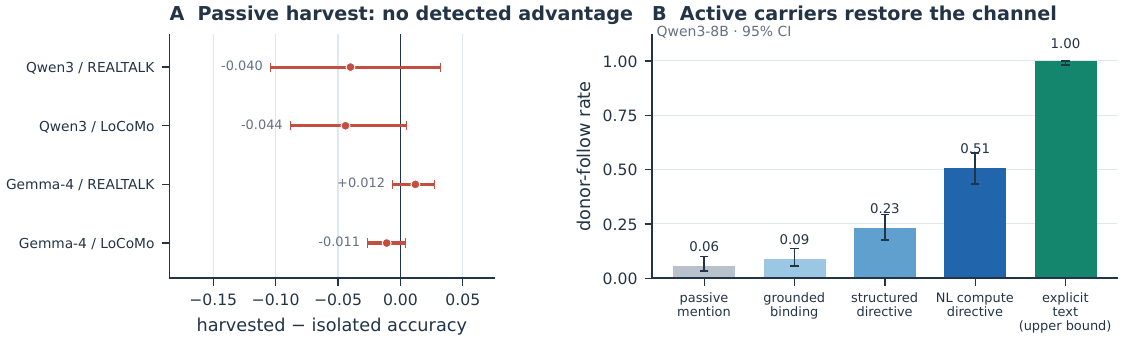}
\caption{\textbf{Passive natural mentions are not a dependable write interface; deliberate
carriers establish the primitive in controlled trajectories.} \textbf{A}: on real long-term
dialogs (REALTALK, LoCoMo), harvesting natural mentions yields no end-to-end benefit over
isolated encoding; Gemma-4 is served natively (original positions, key-masked --- the geometry an
eviction-style system presents, App.~\ref{app:swgate}), where its LoCoMo cell is equivalent to
isolated encoding under the $\pm.05$ band. \textbf{B}: Qwen3-8B donor-follow rate across
the five carrier arms (passive mention to explicit-text upper bound), $95\%$ CIs; the three-recent-model
profiles --- construction-sensitive on Qwen3, saturating on Gemma-4, abstention-dominated on
Ministral-3 under free generation, with explicit text recovering everywhere --- are in
Table~\ref{tab:x9}.}
\label{fig:active}
\end{figure}

\textbf{A deliberate carrier induces donor-aligned recovery.} In donor-paired synthetic trajectories
we emit the carrier ourselves --- one answer-free event after the source, drawn from five arms running
from a passive mention to an explicit-text upper bound. Swapping the passive mention for an
answer-free NL-compute directive (``determine the verdict now and record it here'') lifts
donor-aligned recovery from $6\%$ to $51\%$ on Qwen3 without ever naming the value
(Fig.~\ref{fig:active}B). With the query at a fixed absolute position, the shared downstream review row
alone yields $.00$, while adding the carrier beside it yields $.42$ under the NL directive and $1.00$
under explicit text (Table~\ref{tab:x10}); since merely adding a row \emph{lowers} donor sensitivity
(\S\ref{sec:compose}), the carrier row is what carries. Which construction writes is a per-checkpoint
choice (App.~\ref{app:legacy}), and the one arm that recovers on every recent model is an explicit
textual record. Hence the recipe: \emph{emit compact materialization events for state needed after
its operands leave the serving set, calibrate the construction per checkpoint, and keep an explicit
textual record as the fallback.}

\textbf{Passive harvesting on real dialogs carries nothing of a withheld value.} On REALTALK and
LoCoMo \citep{realtalk2025,locomo2024}, answer-free late mentions harvested from the full-history cache
give no detectable benefit over isolated encodings of the same text (Table~\ref{tab:x8}); on Gemma-4,
served at its original positions, the harvest is equivalent to isolated encoding within $\pm.05$
(LoCoMo $-.011$, $95\%$ CI $[-.026,+.004]$). The null is bounded by effect size rather than by the
instrument: a repaired same-path positive control --- per-item hard negatives, a next-token readout, a
whole-conversation counterfactual and a placebo carrier --- fires, a deliberately pointed carrier
yielding $+.018$ (wild-cluster bootstrap-$t$, $p{=}.040$, $111$ items) as a sparse event. When the
strongest stack of \S\ref{sec:address} selects the served rows with every value-bearing turn withheld
($n{=}1269$), harvested minus isolated is $-.017$ ($p{=}.18$), and $|\delta|$ is bounded within about
$\pm.03$ across a second judge family; co-encoding the served rows is worth $+1.04$ log-odds, but
seeing the rest of the conversation adds nothing at any $k$, and the KV drift a changed fact induces is
diffuse and content-blind (App.~\ref{app:ext-active}, App.~\ref{app:realdialog}).

\textbf{What a harvested row does carry: its own context.} Letting the same stack retrieve freely, the
in-place KV of a served row that states the answer reads it with $\mathbf{+1.87}$ log-odds more
confidence than the same text jointly re-encoded (wild-cluster $p{=}.002$, positive in all ten
conversations), and it follows the row whichever value it states (interaction $+2.97$
$[+2.03,+3.94]$). Accuracy is at ceiling for every arm, so in-place KV is a higher-fidelity
representation of what a row says --- fidelity, not accuracy. The same reading holds on agent
trajectories: on AMA-Bench, replacing every unselected upstream turn with neutral filler changes
$93$--$95\%$ of gathered-KV answers yet moves accuracy by under a point, with flips balanced, for an 8B
and a 32B reader (App.~\ref{app:ama}). Corrections are cheapest as appended patches served by query,
not as cache edits (App.~\ref{app:update}).

\section{Consequences: two deployed heuristics that assume the converse}
\label{sec:consequences}

The contract is directional: a row materializes state that its own computation depended on. Deployed
practice routinely assumes the converse --- that a downstream product, a note or a citation, stands in
for the upstream evidence it was derived from. We test that assumption on $247$ tool-using
trajectories over HotpotQA \citep{hotpotqa2018} with Qwen3-8B: ablation drops a region's cache rows
while every retained row keeps its position, and the readout is the log-odds of a gold value against a
matched decoy, on follow-ups the agent's notes state and on values it read but never wrote down.

\textbf{Prospectively, the measured priority table is the inverse of the hand-set one.} Dropping the
retrieved documents costs $21.99$ nats on unstated follow-ups and dropping any other region costs
approximately nothing, although region-aware schedules give retrieved documents a low priority and a
fast decay \citep{memdecay2026}. The agent's own output carries source values almost only where it
wrote them down ($+1.77$ against $+0.38$), and an answer's KV covers what it stated and no more ($1.00$
on stated facts, $.39$ on unstated ones, against $.97$ for re-retrieval).

\textbf{The attention available at eviction time marks only what need not be kept.} Ranking
$16$-token blocks of the retrieved documents and keeping the top $B\%$, attention accumulated during
the agent's own generation --- the only attention available before the next query arrives --- is useful
on stated follow-ups and indistinguishable from random on unstated ones, while attention from the
follow-up itself shows no such asymmetry (Table~\ref{tab:consequences}). The agent attended to what it
was using to write, so those tokens are the ones whose content already left the cache as text.
Standard eviction validation puts the question in the prompt, which makes accumulated attention
query-aware by construction (App.~\ref{app:ext-consequences}).

\begin{table}[t]
\centering
\caption{\textbf{Whose attention decides whether attention is informative.} Per-item recovery minus
the same item's random baseline, paired over the $n=69$ items with a usable denominator in both
layers; recovery is $1.0$ when keeping the top $B\%$ matches keeping all retrieved text.
Asterisks mark $95\%$ percentile-bootstrap intervals excluding zero ($B=4000$).}
\label{tab:consequences}
\begin{tabular}{llccc}
\toprule
Attention source & Available at eviction & $B$ & Stated & Unstated \\
\midrule
Agent's own generation & \textbf{yes} & $10\%$ & $\mathbf{+0.308^{*}}$ & $+0.061$ \\
Agent's own generation & yes          & $20\%$ & $\mathbf{+0.263^{*}}$ & $+0.045$ \\
The follow-up itself   & no           & $10\%$ & $+0.291^{*}$ & $\mathbf{+0.435^{*}}$ \\
The follow-up itself   & no           & $20\%$ & $+0.343^{*}$ & $\mathbf{+0.501^{*}}$ \\
\bottomrule
\end{tabular}
\end{table}

\textbf{Retrospectively, the reported failure mode is the rarer one.} Run backwards on $418$ cited
multi-hop answers, swapping the value in the answer-bearing passage moves the readout by $+36.1$ nats
(placebo $+0.04$). Decorative citation --- cited but not load-bearing --- accounts for $1.5\%$ of
citations, while $20.7\%$ of load-bearing passages were used and not cited, concentrated on bridge
passages (load-bearing on $76\%$ of items, cited on $30\%$), which an entailment check cannot ask for
\citep{alce2023,wallat2024}.

\textbf{Why dropping the source rarely costs accuracy here.} Drop the tool output in which the answer
first appears and $242$ of the $247$ trajectories still read out the gold value, all of them still
doing so when that output had carried a different value; in $232$ the value survives as text
elsewhere, and the cache channel itself adds $+1.35$ log-odds against $+32.3$ for serving the source.
An unchanged accuracy on real trajectories therefore mostly reflects textual copies; the $99{:}0$ of
\S\ref{sec:discovery} is an existence proof under controlled donor pairs. Scope: one checkpoint and
shallow trajectories (App.~\ref{app:ext-consequences}).

\section{Related work}
\label{sec:related}

\textbf{Closest mechanism work.} Models Take Notes \citep{mtn2026} shows prefill writing memoized
downstream conclusions; MEMENTO \citep{memento2026} is the closest \emph{source-omitted} analogue,
training a model whose memento KV retains information from a masked reasoning block; KVEraser
\citep{kveraser2026} finds that residue persisting after deletion and learns to steer it away; and the
lookback analysis of \citet{lookbacks2025} traces the pointer-style mechanism of in-context state
tracking. We isolate \emph{source-omitted sparse event serving}: unmodified checkpoints, answer-free
spans, and donor-pair identification with served tokens and positions held fixed. \textbf{KV reuse
and repair} --- CacheBlend \citep{cacheblend2024}, EPIC \citep{epic2024}, KEEP \citep{keep2026}, KVLink
\citep{kvlink2025}, HYPIC \citep{hypic2026}, and eviction such as SnapKV \citep{snapkv2024} and H2O
\citep{h2o2023} --- optimizes the physical reuse of the rows it keeps; we audit what an omitted source
left behind. \textbf{Memory systems} --- EpiCache \citep{epicache2025}, Zep \citep{zep2025}, MemGPT
\citep{memgpt2023} --- build the shell; we measure the substrate those shells would need if they
served event KV. App.~\ref{app:ext-related} gives the full comparison.

\section{Limitations and scope}
\label{sec:limits}

We demonstrate a programmable, per-checkpoint sparse-memory primitive, not a universal latent write
API. The donor swaps certify a causal channel for \emph{compact} state: exact numeric payloads are not
read back, derived conclusions need a write-time prompt that itself costs full-context accuracy
($.818\to.698$), and the flat multi-hop curve uses oracle row selection. Representation probes show the
state is present where the readout fails, but decodability is not use. The addressability negative
covers text indexes --- provably the term-frequency family, and on one corpus of ten conversation
clusters the four stages a deployed stack runs --- while a directed graph index failed its gate and no
multi-hop router was tested. The real-dialog null rests on Qwen3-8B and LoCoMo, with wild-cluster
inference at ten clusters. App.~\ref{app:ext-limits} states each limitation in full, including the
serving-path checks, the sliding-window exception, long-context range restoration, and
linear-attention hybrids.

\section{Conclusion}

We omitted the observation an answer depended on and served a downstream event whose text never
states its value. Among donor-sensitive items the answers followed the omitted donor \textbf{99:0}
on Qwen3-8B: an independently served, source-omitted event row can carry a conclusion of
computation whose inputs are gone, so sparse event-KV serving is more than retrieval over rows. The
primitive is \emph{programmable}, and that is the part a system can build on --- a deliberate
answer-free compute carrier lifts donor-aligned recovery from $6\%$ to $51\%$, with an explicit
textual record as the reliable fallback --- inside a contract whose trigger is read through surface form (the row holds the state; the wording decides whether the model says it),
whose landing is the root rather than the edges that point back at it, and whose access is a
compact-state readout envelope. On real long-term dialogs the contract reads directionally: rows a deployed
retrieval stack selects carry nothing of the value it failed to retrieve, under a positive control
that fires, while a row's in-place KV carries its own context with measurably higher fidelity than
a re-encoding of the same text.

One consequence follows for anyone who evicts. An ablation that drops a source event and
observes no accuracy loss has not shown that the source was unnecessary; it may have
retained a row that already carried the answer. Reusable KV can therefore act as
a semantic memory substrate --- under calibrated write and read conditions --- rather than merely
an inference cache: \emph{compute globally, materialize locally, serve sparsely}.

\subsubsection*{Reproducibility}
All experiments run on frozen open-weight models with greedy decoding; donor pairs, leak
assertions, and echo controls are enforced in code. Headline claims rest on three recent
checkpoints --- Qwen3-8B, Ministral-3-8B (instruct), and Gemma-4-12B
\citep{qwen3_2025,ministral3_2025,gemma4_2026} --- each evaluated at one frozen revision in bf16,
so claims are at checkpoint rather than family level; four 2024 checkpoints
\citep{qwen25_2024,gemma2_2024,mistral2023,llama3_2024} appear only as exploratory legacy
diagnostics, and coverage is stated with each result. The release carries the experiment scripts,
an experiment-to-script map, per-script configurations (seeds, arms, readouts), and the exact
repositories, revision SHAs, and environment.\label{lastmainpage}

\bibliography{materialize_refs}
\bibliographystyle{iclr2026_conference}

\appendix

\section{Protocol details}
\label{app:protocols}

\textbf{Serving mechanics.} Events are encoded once per trajectory under the checkpoint's native
causal attention pattern; serving re-assembles selected rows at their original \emph{position ids}
(position-budgeted across donors), so retained keys keep the rotation they were written with, and
appends the query fresh. The rows are concatenated into contiguous cache \emph{slots}: for
full-attention layers this is immaterial, but where attention is windowed the window is counted in
slots, which is what App.~\ref{app:swgate} measures. Isolated controls re-encode the identical
served text without the trajectory prefix. Greedy decoding throughout; readouts are
short-answer with menu-in-question where noted; forced-choice logit probes calibrate
on text arms and are disclosed as output-interface (not representation) probes.

\textbf{Statistics.} Wilson $95\%$ intervals for rates; McNemar-style discordant counts for
paired arms; conversation/trajectory-clustered bootstrap for dialog data. Equivalence to zero
uses two one-sided tests (TOST): the effect is \emph{equivalent} only if its $90\%$ interval lies
entirely within a pre-registered bound; directional effects use $95\%$ intervals.
Exact sign-test $p$-values are computed without floating-point tolerance, and extreme-tail values
are independently re-derived (e.g.\ the 99:0 value is $3.2{\times}10^{-30}$). The headline causal effects
replicate with near-perfect directionality on 2026 Gemma-4
(\S\ref{sec:discovery},~\S\ref{sec:compose}). For the real-dialog follow-up
(App.~\ref{app:realdialog}), whose clusters are ten to seventeen conversations, intervals and
$p$-values are wild-cluster bootstrap-$t$ (Rademacher weights, null imposed, all $2^{G}$ sign
vectors enumerated, so no Monte-Carlo error); a synthetic size check on the actual cluster
structure found the percentile cluster bootstrap rejecting $11\%$ of true nulls at nominal $5\%$
while the wild-cluster test held $5\%$, so the pre-registered percentile intervals were replaced
before any effect was read.

\textbf{Donor pairs, the operative null, and transition classes.} Each item builds two
trajectories differing only in the omitted source (e.g.\ a register set \textsc{online} vs.\
\textsc{offline}; a sensor reading above vs.\ below a threshold). Under the byte-identity control
with greedy decoding, the two donor runs would produce \emph{identical} answers absent any
contextual carryover, so donor dependence is established whenever the two decoded outputs
\emph{differ}. We partition each pair into \textsf{follow} (donor-aligned --- each donor answers
its own gold), \textsf{anti} (anti-aligned), \textsf{const} (identical committed answer), and
\textsf{other} (the tables' residual: identical abstentions, which are donor-\emph{blind}, together
with discordant unscorable pairs where exactly one side abstains). Only \textsf{follow},
\textsf{anti}, and the discordant subset of \textsf{other} have differing outputs and thus witness
donor dependence; \textsf{const} and identical abstentions do not. The exact two-sided sign test on
\textsf{follow}:\textsf{anti} is then a separate, conditional test of the \emph{direction} of that
dependence, not a test of the causal null itself. Sources are
position-budgeted so both donor variants occupy identical position ranges; all served spans
are asserted byte-identical across donors in latent arms; leakage assertions reject any
carrier containing answer tokens or operands (one of our own templates was rejected by this
assertion during the smoke run and rephrased).

\textbf{Recent-model behavioral-compatibility gates and long-context mechanics.} Each recent
2025--26 model passed a pre-experiment compatibility gate --- behavioral agreement, not bit-level
identity --- between one-shot text prefill and encode--assemble--continue (greedy), reported at
three levels (content, first-token, logit deviation).
Qwen3 agrees with one-shot prefill on $21/24$ full decoded answers (both arms
$24/24$ task-correct, so the three discrepancies are wording, not content) and on $23/24$
first-token argmax over identical ids; the lone first-token miss is a bf16 near-tie --- the
one-shot top-1/top-2 margin there is $0.25$, below the $0.53$ path deviation (max$|\Delta\text{logit}|$
mean $0.59$) --- and a header-consistency control is $24/24$. The piecewise-vs-one-shot
\emph{tokenization} mismatch is a BPE-boundary artifact, which is why the gate is scored on
identical ids. The two models that needed non-standard handling: Gemma-4 23/24
exact-string agreement with 24/24 correctness on both arms;
Ministral-3 21/24 exact with 24/24 content agreement on both arms --- its three
divergences are answer-preserving template-choice flips on a verbose model, and the
content-level criterion for this case is documented here rather than silently applied.
Long-context encodes on Gemma-4 use causally-chunked prefill (4k chunks through the
same cache; algorithmically equivalent under the model's causal and sliding-window masks, and
verified against one-shot prefill at lengths where one-shot execution was feasible), needed because fused
kernels reject that model's custom attention mask at 30--40k tokens. The donor
factorial and true-edge experiments (\S\ref{sec:discovery},~\S\ref{sec:compose}) are
read out on Gemma-4 with a menu-form direct question rather than the trace CoT used for
Qwen3, because Gemma-4's free-generation interface is unreliable (\S\ref{sec:write}); a
same-protocol Qwen3 control reproduces its causality in the \emph{mention-absent root} cell
($88{:}0$; the true-edge run's own root-only bridge cell gives $89{:}0$) and on the edge
($84{:}0$). Under this protocol Qwen3's \emph{co-served} cells saturate to a single answer
(root contrast $1{:}0$), so this control speaks to the root-only drive, not the co-served
factorial.

\textbf{Active-materialization arms (\S\ref{sec:active}).} $192$ donor pairs per arm per
model, two templates per arm, 24-event trajectories, source at mid-trajectory, carrier
in the following slot; served set = \{decoy, unrelated register, carrier, downstream review\};
unrelated-query echo control served in all arms. A secondary target-addressed readout
(``what is recorded in register $T$'') moves the rates in \emph{opposite} directions by
construction on Qwen3 --- grounded binding rises from $.089$ to $.188$ while the structured
directive falls from $.229$ to $.120$ and explicit text from $1.00$ to $.885$ --- which is the
query-form dependence of \S\ref{sec:envelope} and why the preferred address is a property of the
construction rather than of the model; legacy-model target-addressed rates are in
Appendix~\ref{app:legacy}.

\textbf{Real-dialog protocol (Fig.~\ref{fig:active}A).} Qualifying QA require at least
one answer-free late mention after the evidence session; candidate order balanced by
item; harvested vs.\ isolated encodings compared paired per question over the full
qualifying set (no strong-entity subsetting), conversation-clustered bootstrap
(seed-pinned, $B{=}4000$); equivalence band $\pm.05$ pre-registered. A
gold-evidence arm (serving the official evidence turns) certifies the tasks are answerable ($.85$--$.96$
under the two-choice (gold vs.\ hard-negative) recognition readout, well above its $.5$ chance baseline);
recognition readout removes the free-generation bottleneck documented in \S\ref{sec:envelope}.

\section{Additional tables}
\label{app:tables}

\textbf{Factorial estimator and ITT sensitivity (Table~\ref{tab:factorial}).} Each base item contributes one paired contrast (root $=$ the mean of the two root differences at fixed mention; mention likewise; interaction $=$ their difference); intervals are normal-approximation (paired-Wald) over those per-item values, so clustering is at the item. Complete-case exclusion is outcome-conditioned, so we re-ran every factorial effect counting non-\{ONLINE,OFFLINE\} readouts as non-ONLINE over \emph{all} items: nothing moves by more than $.01$ (Qwen3 root $+.082$, mention $-.027$, TOST $p{=}.038$; Gemma-4 root $+.211$, mention $+.159$). The mention-absent column comes from its own serving protocol and is re-analysed separately: on Qwen3 it is unchanged at $+.398$ (no non-binary readouts there), and on Gemma-4 it moves from $+.091$ to $+.109$.

\begin{table}[h]
\centering
\caption{Four-cell counts for the source-omitted 2$\times$2 donor factorial
(Table~\ref{tab:factorial}): full per-cell \textbf{ONLINE/OFFLINE/other} tallies over
\emph{all} items ($n{=}128$ Qwen3, $192$ Gemma-4), so the \textsf{other} (non-binary) readouts
are visible rather than conditioned away; the effects in Table~\ref{tab:factorial} use the
complete-case subsets ($n{=}126$/$182$), with an ITT re-analysis reported there. The last two
columns are the mention-absent cells; note their root contrast is measured with the query
closer to the root (App.~\ref{app:poscontrol}). Root donor sets the queried register's state; the mention donor is the value-free
downstream reference. Qwen3-8B free-generation readout; Gemma-4-12B menu readout.}
\label{tab:factorial_cells}
\footnotesize
\setlength{\tabcolsep}{4pt}
\begin{tabular}{lcccccc}
\toprule
& \multicolumn{2}{c}{root${=}$ON} & \multicolumn{2}{c}{root${=}$OFF} & \multicolumn{2}{c}{mention absent} \\
Checkpoint & men${=}$ON & men${=}$OFF & men${=}$ON & men${=}$OFF & root${=}$ON & root${=}$OFF \\
\midrule
Qwen3-8B    & 64/63/1 & 67/60/1 & 53/74/1 & 57/71/0 & 82/46/0 & 31/97/0 \\
Gemma-4-12B & 121/63/8 & 89/98/5 & 79/109/4 & 50/139/3 & 103/88/1 & 82/106/4 \\
\bottomrule
\end{tabular}
\end{table}

\begin{table}[h]
\centering
\caption{X8 same-path injected-carrier positive control (\S\ref{sec:active}): gold recovery
with an injected donor-paired carrier vs.\ isolated encoding of the same carrier text, routed
through the identical harvest/splice/recognition path ($n{=}62$ QA each, REALTALK$+$LoCoMo).
Both rows use the \emph{identical} $62$-item set over $17$ conversations. Both checkpoints are
served here through the compact path, so for Gemma-4 this control speaks to the
splice/recognition path rather than to the native geometry of Table~\ref{tab:x8}. Paired difference
with McNemar exact $p$ and discordants (injection-helps:injection-hurts), plus a
conversation-clustered bootstrap $95\%$ CI (seed-pinned $B{=}4000$, matching
Table~\ref{tab:x8}); the McNemar $p$ is item-level, and clustering changes neither
conclusion. Qwen3's forced choice is option-prior-saturated (a counterfactual carrier still
selects gold ${\sim}80\%$), so its control cannot fire under this frozen protocol;
App.~\ref{app:realdialog} repairs the option prior, the readout and the counterfactual, after which
the Qwen3 control fires ($+.018$, interval excluding zero; Table~\ref{tab:e3}). Gemma-4 \emph{does} separate
($+.194$, clustered CI excludes zero): a detected end-to-end contextual benefit, whose $+12$
gold gain is exactly matched by twelve fewer abstentions (\textsf{None} $42\to30$; wrong
unchanged at $3$). The donor-flipped counterfactual arm was collected but fires on too few
items to adjudicate --- Gemma-4 $3{:}0$ follow:anti with $33/62$ counterfactual readouts
unscorable, Qwen3 $1{:}0$ --- so this does not separate state-specific transfer from generic
readout activation, and we report it as a \emph{partial} positive.}
\label{tab:x8ctrl}
\small
\begin{tabular}{lccccc}
\toprule
Checkpoint & inj.\ gold & iso.\ gold & $\Delta$ & McNemar $p$ (disc.) & clustered $95\%$ CI \\
\midrule
Qwen3-8B    & 50/62 & 51/62 & $-.016$ & $1.00$ ($3{:}4$)  & $[-.102,+.067]$ \\
Gemma-4-12B & 29/62 & 17/62 & $+.194$ & $.043$ ($21{:}9$) & $[+.030,+.357]$ \\
\bottomrule
\end{tabular}
\end{table}

\textbf{Construction classes.} The recent-model write bank is Table~\ref{tab:bank}; Gemma-4
reaches $\ge.90$ candidate-logit reachability on 8/16 and $\ge.85$ on 10/16 under the trace
protocol despite writing none. The four legacy 2024
models (Gemma-2 8/1/5/2; Qwen2.5 1/6/9/0; Llama-3.1 0/2/14/0; Mistral 0/1/2/13,
comprehension-failed and excluded from write claims) are in Appendix~\ref{app:legacy}.
The trace and direct readouts are not a fixed offset: on Gemma-4 they differ by $-.46$ to $+.65$
across the sixteen constructions, with the trace rate the higher of the two on eight of them, so
which readout a construction favours is itself construction-specific (protocol constant across
cells; cross-section comparisons carry the protocol).

\textbf{Two orthogonal bits over the four-way payload.} A four-way readout at chance does not by
itself show the rows hold less than two bits --- four-way \emph{decoding} could be the bottleneck.
On the same served cache we therefore also ask Qwen3 two binary questions whose answers are orthogonal
halves of the same variable (``if \textsc{north} or \textsc{south} answer \textsc{red}, else
\textsc{blue}'', and the diagonal split). Nothing in the cell beats answering a constant: over
$n{=}256$ the realised label split makes the best constant answer worth $.559$ and $.512$ on the two
bits and $.281$ on the four-way question, against observed $.500$ $[.439,.561]$, $.508$
$[.447,.568]$ and $.285$ $[.233,.343]$, with the bit pair reconstructing the four-way answer at
$.262$ $[.212,.319]$. Nor is the model merely declining to emit a four-way label: it answers the
binary halves with a strong bias of its own ($189/256$ \textsc{blue} on the first) and still lands
on the floor. The envelope of \S\ref{sec:envelope} is therefore not an artifact of four-way decoding; it is a
statement about what the model's readout recovers from the rows --- representation probes recover a four-way value from the served values of matched payload items at $\approx\!1.00$ (App.~\ref{app:repprobe}). Labelling the two questions
\textsc{yes}/\textsc{no} instead is uninformative --- the model answers \textsc{no} almost always
($.047$ and $.004$ \textsc{yes}) --- which is why the neutral pair is used.

\textbf{Where in a mirror chain the value sits.} A depth-3 chain --- root $C_1$ (``$M_1$ mirrors
$S$''), then $C_2$ (``$M_2$ mirrors $M_1$''), then $C_3$ (``$T$ mirrors $M_2$''), source dropped ---
separates a value re-resolved into every link from one that lives in the root alone. Served link by
link, only the root answers: $C_1$ alone $.699$ $[.640,.752]$, against $C_2$ alone $.465$ and $C_3$
alone $.484$ --- level with the $.473$ text floor and below the $.539$ a constant answer earns on
this label split ($n{=}256$, binary menu). Dropping the root and serving
$\{C_2,C_3\}$ leaves $.465$; adding it back restores $.707$ (McNemar $68{:}6$,
$p{=}2{\times}10^{-14}$). The chain is read by walking the served text back to the root and taking
the value from the root's rows, which is what \S\ref{sec:compose}'s ``roots carry, edges route''
amounts to mechanically --- and it is the contrast that separates that reading from the alternative
in which each link re-resolves the value into rows of its own.

\textbf{Source-present multi-hop denoising ceiling.} When the
source \emph{is} served, selecting the dependency-chain rows holds accuracy nearly flat across
depth --- $128/128$, $128/128$ and $121/128$ at $1$, $2$ and $4$ hops ($377/384$ overall) ---
while full-context reading collapses on far spans
(16/384, $p{\approx}4{\times}10^{-107}$). A four-arm decomposition attributes this to
routing/denoising: clean text re-encoding of the same selected rows performs comparably
(384/384 vs.\ 377/384), so the gain is not source-omitted materialization. Selection is oracle (mechanism ceiling,
not a deployable number); we report it only to separate it from the source-omitted
channel of \S\ref{sec:compose}.

\textbf{Real-dialog passive-harvest audit.} Table~\ref{tab:x8} gives the authoritative
per-cell numbers for Fig.~\ref{fig:active}A: qualifying-QA count, conversation clusters,
harvested$-$isolated point estimate, and $90/95\%$ conversation-clustered intervals.

\begin{table}[h]
\centering
\caption{Real-dialog passive harvest (harvested $-$ isolated recognition accuracy;
per-question paired, conversation-clustered bootstrap, seed-pinned $B{=}4000$; $n$ =
qualifying QA, clu = conversation clusters). Designated equivalence to $0$ uses a $\pm.05$
TOST on the $90\%$ interval. Both checkpoints are evaluated on the \emph{identical}
qualifying sets ($497$ REALTALK / $564$ LoCoMo QA over $10$ conversations each). Qwen3 shows no
benefit on REALTALK and a marginal negative on LoCoMo ($90\%$ excludes $0$, $95\%$ does not), and
reaches $\pm.05$ equivalence on neither --- so we log ``no advantage detected,'' not equivalence.
The Gemma-4 rows are served \emph{natively} --- retained rows at their original
positions, omitted rows key-masked, query at the original end position --- which is the geometry an
eviction-style serving system presents (App.~\ref{app:swgate}); they are read by candidate logit,
as elsewhere in this paper. Under that geometry Gemma-4 harvesting on LoCoMo is \emph{equivalent} to
isolated encoding within the $\pm.05$ band --- the only cell here attaining designated equivalence
rather than an undetermined null. $^{\dagger}$REALTALK's ten dyads form three participant
components, so its intervals are descriptive and support no cluster-based inference; formal
verdicts are LoCoMo-only. A legacy Gemma-2 replication (negative) is in
Appendix~\ref{app:legacy}.}
\label{tab:x8}
\footnotesize
\setlength{\tabcolsep}{3.2pt}
\begin{tabular}{llccccl}
\toprule
Model & Dataset & $n$/clu & harv$-$iso & 90\% CI & 95\% CI & verdict \\
\midrule
Qwen3-8B    & REALTALK$^{\dagger}$ & 497/10 & $-.040$ & $[-.094,+.020]$ & $[-.104,+.032]$ & descr. \\
Qwen3-8B    & LoCoMo   & 564/10 & $-.044$ & $[-.081,-.005]$ & $[-.088,+.005]$ & inconcl. \\
\addlinespace
Gemma-4-12B & REALTALK$^{\dagger}$ & 497/10 & $+.012$ & $[-.002,+.025]$ & $[-.006,+.027]$ & descr. \\
Gemma-4-12B & LoCoMo   & 564/10 & $-.011$ & $[-.024,+.002]$ & $[-.026,+.004]$ & equiv. \\
\bottomrule
\end{tabular}
\end{table}

\textbf{The contract across models (consolidated).} A consolidated cross-model
panel including the legacy 2024 models (Table~\ref{tab:gen}) is in
Appendix~\ref{app:legacy}; the tested Qwen3.6-27B is out of scope by substrate (\S\ref{sec:limits}).

\begin{table}[h]
\centering
\caption{\textbf{An answer-free compute directive lifts donor-aligned recovery from $11/192$ under a passive mention to $97/192$ on Qwen3-8B.} X9 donor-transition classes on the subject-addressed readout under harvested
serving (follow/anti/const/other; $192$ donor pairs per arm per model). \arm{exp} carries the answer in text (upper bound). Three
recent-model profiles: \emph{Qwen3} is construction-sensitive with a non-saturating
latent positive (\arm{dir}$-$\arm{pas} $+.172$ CI $[.109,.234]$, $p{=}2.5{\times}10^{-7}$;
\arm{nld}$-$\arm{pas} $+.448$, $p{=}5.8{\times}10^{-25}$); \emph{Gemma-4} materializes every
construction to ceiling; \emph{Ministral-3} is abstention-dominated in every latent arm
(no reliable donor-aligned advantage), recovering only explicit text.
\textbf{Readouts.} They follow each model's free-generation reliability: Qwen3 and
Ministral-3 via free-generation trace, Gemma-4 via the menu readout it requires
(\S\ref{sec:write}). Legacy
2024 columns (Gemma-2, Qwen2.5) are in Appendix~\ref{app:legacy}.}
\label{tab:x9}
\small
\begin{tabular}{lccc}
\toprule
Arm & Qwen3-8B & Gemma-4-12B & Ministral-3-8B \\
 & \multicolumn{3}{c}{\textsf{follow}/\textsf{anti}/\textsf{const}/\textsf{other} of $192$} \\
\midrule
\arm{pas} passive mention      & 11/0/181/0  & 189/0/3/0   & 0/0/4/188  \\
\arm{bind} grounded binding    & 17/0/175/0  & 191/0/1/0   & 0/0/0/192  \\
\arm{dir} structured directive & 44/0/148/0  & 191/0/1/0   & 6/3/64/119 \\
\arm{nld} NL compute directive & 97/0/95/0   & 188/0/4/0   & 4/4/71/113 \\
\arm{exp} explicit text        & 192/0/0/0   & 192/0/0/0   & 182/0/0/10 \\
\bottomrule
\end{tabular}
\end{table}

\begin{table}[h]
\centering
\caption{X10 serve-set ablation on the \emph{three recent models} (Qwen3, Gemma-4,
Ministral-3 --- three families; donor-follow rate serving the carrier row alone, the shared
downstream-review row alone, or both --- \textbf{c}/\textbf{r}/\textbf{b}; $192$ donor pairs
per serve-set cell per model; same manifest/seeds/menu readout; retained decoy and
unrelated-register rows precede the source and are donor-invariant; full
\textsf{follow}/\textsf{anti}/\textsf{const}/\textsf{other} counts in
Table~\ref{tab:x10full}). Retained rows keep their original absolute positions; the query is
appended after the last served row, so --- as the review is always the trajectory's final
row --- \textbf{r} and \textbf{b} share the query position while \textbf{c} places it
earlier. On these checkpoints the carrier carries the dominant donor-aligned signal: under
the query-position-controlled contrast \textbf{r}$\to$\textbf{b}, adding the carrier drives
Qwen3 follows $0\to.42$ (\arm{nld}; review alone shows no detected effect), and Gemma-4 saturates the carrier on
every arm. The
\textbf{c}-vs-\textbf{b} contrast also moves the query, so the co-served-review change
(\arm{nld} $.19\to.42$) is an interface-level serve-set effect, not a span interaction.
Gemma-4 keeps a weak \emph{donor-aligned} review-local signal (review-only
\arm{nld}/\arm{exp} $18{:}0$/$15{:}0$ follow:anti); Ministral-3 is abstention-dominated on
latent arms, its explicit-text review row is donor-sensitive but \emph{anti}-aligned
($0{:}18$; Table~\ref{tab:x10full}) --- not inert. Ministral-3's passive-carrier drop under
co-serving (\arm{pas} $.09\to.00$) is \emph{not} identified: the query-controlled
\textbf{r}$\to$\textbf{b} contrast shows no detected change ($1\to0$, $p{=}1.0$), so the cause
(query offset vs.\ review membership) is unresolved. A \emph{review-dominant} shift, the review
\emph{leading}, appears only in the exploratory Gemma-2 run (App.~\ref{app:legacy}) and
is not reproduced on any recent model.}
\label{tab:x10}
\footnotesize
\setlength{\tabcolsep}{2.6pt}
\begin{tabular}{l|ccc|ccc|ccc}
\toprule
& \multicolumn{3}{c|}{Qwen3-8B} & \multicolumn{3}{c|}{Gemma-4-12B} & \multicolumn{3}{c}{Ministral-3-8B} \\
Arm & c & r & b & c & r & b & c & r & b \\
\midrule
\arm{pas}  & .00 & .00 & .03 & .98 & .01 & .99 & .09 & .01 & .00 \\
\arm{bind} & .06 & .00 & .09 & .99 & .04 & .99 & .00 & .02 & .00 \\
\arm{dir}  & .21 & .00 & .25 & 1.00 & .07 & 1.00 & .00 & .04 & .03 \\
\arm{nld}  & .19 & .00 & .42 & .99 & .09 & .98 & .00 & .02 & .02 \\
\arm{exp}  & .97 & .00 & 1.00 & 1.00 & .08 & 1.00 & .98 & .00 & .98 \\
\bottomrule
\end{tabular}
\end{table}

\begin{table}[h]
\centering
\caption{X10 full transition counts (\textsf{follow}/\textsf{anti}/\textsf{const}/\textsf{other};
$192$ donor pairs per cell) for the three recent models, serving carrier-alone / review-alone /
both. \textsf{const} = donor-blind committed answer; \textsf{other} = at least one unscorable donor
output --- a donor-blind identical abstention, or a discordant pair where exactly one side abstains
(donor-\emph{dependent} but undirected). The
interpretable, query-position-controlled contrast is \textbf{review}$\to$\textbf{both} (query
fixed at the trajectory's final review row): adding the carrier drives Qwen3 \arm{nld}
follow $0\to81$ (McNemar $81{:}0$, exact $p{=}8{\times}10^{-25}$; \arm{dir} $0\to48$),
replicated $0\to98$ under a disjoint seed ($n{=}192$; maximum per-cell follow-rate deviation
$.09$/$.03$/$.05$ across the three models). Gemma-4's review-only keeps a weak donor-aligned
signal (\arm{nld} $18{:}0$) and Ministral-3's explicit-text review is \emph{anti}-aligned
($0{:}18$, replicated $2{:}21$).}
\label{tab:x10full}
\footnotesize
\setlength{\tabcolsep}{3pt}
\begin{tabular}{llccc}
\toprule
Model & Arm & carrier & review & both \\
\midrule
\multirow{5}{*}{Qwen3-8B}
 & \arm{pas}  & 0/0/192/0   & 0/0/192/0   & 5/0/187/0   \\
 & \arm{bind} & 12/0/180/0  & 0/0/192/0   & 17/0/175/0  \\
 & \arm{dir}  & 41/0/151/0  & 0/0/192/0   & 48/0/144/0  \\
 & \arm{nld}  & 36/0/156/0  & 0/0/192/0   & 81/0/111/0  \\
 & \arm{exp}  & 187/0/5/0   & 0/0/192/0   & 192/0/0/0   \\
\midrule
\multirow{5}{*}{Gemma-4-12B}
 & \arm{pas}  & 188/0/4/0   & 2/0/166/24  & 191/0/1/0   \\
 & \arm{bind} & 190/0/2/0   & 7/0/155/30  & 191/0/1/0   \\
 & \arm{dir}  & 192/0/0/0   & 13/0/169/10 & 192/0/0/0   \\
 & \arm{nld}  & 191/0/1/0   & 18/0/58/116 & 189/0/3/0   \\
 & \arm{exp}  & 192/0/0/0   & 15/0/154/23 & 192/0/0/0   \\
\midrule
\multirow{5}{*}{Ministral-3-8B}
 & \arm{pas}  & 18/0/17/157 & 1/2/46/143  & 0/1/6/185   \\
 & \arm{bind} & 0/0/20/172  & 4/4/46/138  & 0/0/0/192   \\
 & \arm{dir}  & 0/0/22/170  & 7/8/115/62  & 5/8/67/112  \\
 & \arm{nld}  & 0/1/2/189   & 3/4/83/102  & 3/4/80/105  \\
 & \arm{exp}  & 189/0/0/3   & 0/18/36/138 & 188/0/0/4   \\
\bottomrule
\end{tabular}
\end{table}

\textbf{X9 transition classes.} Table~\ref{tab:x9} gives the full counts on the three
recent models. Unrelated-query flips $0/192$ with accuracy $1.00$ in every arm --- the
materialization channel does not echo into unrelated readouts; \textsf{anti}$\le 4/192$
throughout. On Qwen3 and Gemma-4 the non-follow remainder is \textsf{const}; Ministral-3's
latent arms are dominated by \textsf{other} ($113$--$192/192$ end without a verdict word
under free generation --- verbose deliberation, ``To answer this question, we need to
analyze the trajectory\dots''), the free-generation-interface property behind its
abstention profile (as with legacy Qwen2.5, App.~\ref{app:legacy}). Isolated floors are
donor-blind by byte-identity.

\section{Legacy-model diagnostics}
\label{app:legacy}

The four 2024-generation models (Gemma-2-9B, Qwen2.5-7B, Llama-3.1-8B, Mistral-7B-v0.3)
were run as \emph{exploratory} mechanism diagnostics with uneven coverage: all four ran the
construction bank, Gemma-2 and Qwen2.5 additionally ran the active-materialization arms
(X9), and Gemma-2 alone ran the serve-set ablation (X10) and the true-edge experiment
(its $174{:}0$ cell in Table~\ref{tab:gen}). They support none of
the headline claims, which rest on the three recent models
(\S\ref{sec:discovery}--\S\ref{sec:active}); we report them here for transparency and
because two mechanism observations are currently sharpest on a legacy model.
Model selection was not frozen against these outcomes, so we do not use the legacy
panel to assert any population-level regularity --- only to widen the aperture range
descriptively and to disclose a stability boundary.

\begin{figure}[h]
\centering
\includegraphics[width=\linewidth]{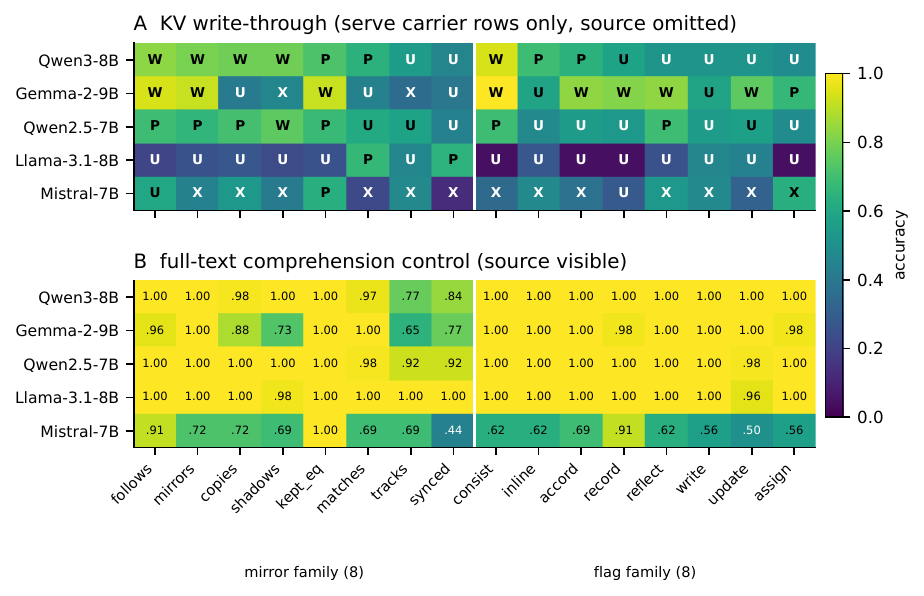}
\caption{\textbf{Cross-generation construction bank} (exploratory):
Qwen3 as a recent anchor and four 2024 legacy models. Sixteen constructions $\times$ five
models (Qwen3/Gemma-2/Qwen2.5/Llama/Mistral). \textbf{A}: KV write-through (serve carrier rows
only, source omitted), class W/P/U/X. \textbf{B}: full text visible --- near-uniformly
solved except Mistral. $n$ per cell: 64/48/64/48/32.}
\label{fig:write_legacy}
\end{figure}

\textbf{Construction bank (Fig.~\ref{fig:write_legacy}).} Across the five-model sweep,
Gemma-2 has the widest trace-readout aperture (8/16 W), Llama the narrowest (0/16 W, 14/16 U
with a majority of KV-arm readouts silent), and Mistral-7B fails the comprehension control itself
(13/16 X, excluded from write claims). The $U_{\text{logit}+}$ split is visible here too:
Llama \texttt{consist} probe $.72$ CI $[.57,.84]$ ($29/40$) vs.\ generation $.10$; Gemma-2
\texttt{accord}/\texttt{inline}/\texttt{reflect} probe $1.00$ vs.\ generation $.38$--$.69$.

\begin{table}[h]
\centering
\caption{Selected cross-model panel (recent + legacy; Qwen2.5 omitted for space --- it ran
the construction bank and X9). Bank~W = W-class
constructions under the trace protocol (of 16); probe = candidate-logit reachability;
edge = follow:anti among donor-sensitive pairs, ask-through-edge, and the parenthetical is
the \emph{same run's} root-only bridge cell (serve \{decoy, root\}, ask the mirror register)
under the same readout ($^\dagger$ = menu readout, which Gemma-4's free-generation interface
requires; unmarked cells free-generation). The \S\ref{sec:discovery} headline $99{:}0$ is a
separate root-only run under free generation, not this run's root cell. X9 latent = best
latent-carrier donor-follow rate (abst.\ = abstention-dominated, no reliable latent positive);
X9 expl.\ = explicit-text follow rate; X10 serve-set results are in Table~\ref{tab:x10}. Legacy
rows (2024) are exploratory.}
\label{tab:gen}
\small
\setlength{\tabcolsep}{3pt}
\begin{tabular}{llccccc}
\toprule
Model & Year & Bank W & Probe & Edge (root) & X9 latent & X9 expl. \\
\midrule
Mistral-7B     & 2024 & excl.\ (13 X) & --- & --- & --- & --- \\
Gemma-2-9B     & 2024 & 8/16  & 8/16 $\ge.90$ & 174:0 & .53 & 1.00 \\
Llama-3.1-8B   & 2024 & 0/16  & split & --- & --- & --- \\
Qwen3-8B       & 2025 & 5/16  & $\approx$gen & 84:0$^\dagger$ (root 89:0$^\dagger$) & .51 & 1.00 \\
Ministral-3-8B & 2025 & 7/16  & --- & \textbf{298:0} & abst. & .95 \\
Gemma-4-12B    & 2026 & 0/16 & $\ge.90$ on 8/16 & 41:0$^\dagger$ & $\approx$1.0 & 1.00 \\
\bottomrule
\end{tabular}
\end{table}

\textbf{Active materialization (legacy).} Table~\ref{tab:x9leg} gives the legacy counts.
On Gemma-2 the deliberate-carrier ordering \emph{reverses} between the X9 free-generation
run (\arm{nld} $.53$ > \arm{pas} $.44$) and the extended-readout serve-set ablation
(\arm{pas} $.55$ > \arm{nld} $.31$): the \arm{nld} cell falls $.53\to.31$ while the
passive$\to$\arm{nld} ordering inverts --- a template/manifest/readout-protocol sensitivity
we flag as a stability boundary. The
serve-set ablation also localizes the signal to the \emph{review} row for grounded-binding
and the structured directive (carrier $.03$/$.01$ vs.\ review $.46$/$.42$) --- the sharpest
instance of \emph{trigger $\neq$ landing span}, and evidence the served set can shift the
readout in either direction (\arm{nld} both $.31$ $<$ carrier $.44$; \arm{bind} both
$.09$ $<$ review $.46$; like the recent-model serve-set contrasts, these move the query as
well as the row set, \S\ref{sec:active}). The \emph{review-dominant} shift is \emph{observed only in the exploratory Gemma-2 run} among checkpoints with X10 coverage: on the recent models the
review row never leads the carrier \emph{in donor-aligned signal} (Ministral-3's abstaining
review arms carry a few raw follows above its floored carrier, but with matching
\textsf{anti}). The review row is not silent on all of them, though ---
Gemma-4's review carries a weak donor-aligned signal (\arm{nld} $18{:}0$, \arm{exp}
$15{:}0$), and Ministral-3's explicit-text review is donor-sensitive but \emph{anti}-aligned
($0{:}18$), while Qwen3's review stays at floor (Table~\ref{tab:x10full}) --- but none is a
review-dominant landing, so the shift did not reproduce on any recent model.
Target-addressed readout collapses on Gemma-2's latent arms ($.026$ binding, $.000$ directive,
against $1.00$ for explicit text) --- a source-keyed latent-note boundary, and the opposite of the
construction-dependent split Qwen3 shows (App.~\ref{app:protocols}). Qwen2.5 abstains in every latent arm at both 12- and 48-token
budgets. The pre-registered ``directive $>$ passive on both models'' criterion
\emph{fails} on Gemma-2 (directive $.000$ vs.\ passive $.443$). On the real-dialog
passive-harvest audit (Fig.~\ref{fig:active}A), the legacy Gemma-2 cells are also negative
(REALTALK $-.034$, $95\%$ CI $[-.069,-.002]$; LoCoMo $-.035$, $[-.063,-.010]$;
conversation-clustered), consistent with the recent models.

\begin{table}[h]
\centering
\caption{Legacy active-materialization counts (exploratory; supports no headline). Left:
X9 donor-transition (follow/anti/const/other, 192 pairs/arm). Right: Gemma-2 serve-set
ablation (carrier/review/both donor-follow rate).}
\label{tab:x9leg}
\small
\begin{tabular}{l|cc|ccc}
\toprule
& \multicolumn{2}{c|}{X9 counts} & \multicolumn{3}{c}{Gemma-2 X10} \\
Arm & Gemma-2 & Qwen2.5 & c & r & b \\
\midrule
\arm{pas}  & 85/0/107/0 & 0/0/61/131 & .32 & .24 & .55 \\
\arm{bind} & 14/0/178/0 & 0/1/21/170 & .03 & \textbf{.46} & .09 \\
\arm{dir}  & 0/0/192/0  & 0/0/46/146 & .01 & \textbf{.42} & .01 \\
\arm{nld}  & 101/0/88/3 & 0/0/74/118 & \textbf{.44} & .09 & .31 \\
\arm{exp}  & 192/0/0/0  & 190/0/0/2  & 1.00 & 1.00 & 1.00 \\
\bottomrule
\end{tabular}
\end{table}

\section{Sliding-window oracle gate (Gemma-4)}
\label{app:swgate}

Gemma-4-12B interleaves local sliding-window and global attention ($40$ of $48$ layers are
sliding, window $1024$). Our compact serving (\S\ref{app:protocols}) re-assembles retained rows at
contiguous cache slots and appends the query at compact \texttt{cache\_position}; the local-attention
window is therefore measured in \emph{compact} slots, not original positions. We test whether this
reproduces native serving. For a donor pair whose \emph{far} row directly states the queried
register's value (position-budget padded so the two donor variants are byte-identical outside the
value), we serve $\{$far row, near decoy$\}$ two ways from one shared prefill --- so only the query's
attention geometry differs: \textbf{compact} (\texttt{assemble}+greedy, the deployed path) versus
\textbf{native} (retained rows kept at original positions in the full cache, the model's native
sliding-window mask, plus a key-mask dropping the omitted rows). Readout is candidate-logit
(the menu protocol Gemma-4 requires); $n{=}32$ per gap. Table~\ref{tab:swgate} reports both paths
as the far row's original distance crosses the window.

\begin{table}[h]
\centering
\caption{Compact vs.\ native serving on Gemma-4 as the far row's original distance from the query
crosses the $1024$ window. \emph{gap} = query-to-far-row distance in tokens. Within one window the
two paths agree (aggregate follow counts identical; first-token argmax $29/32$ on the ONLINE-donor
arm); beyond it they diverge sharply.
The compact path does not simply over-expose the far row --- its position/slot discontinuity
\emph{attenuates} it --- but the served numbers are not native.}
\label{tab:swgate}
\small
\begin{tabular}{lcccc}
\toprule
gap (tok) & window & native follow:const & compact follow:const & argmax agree; max$|\Delta\text{logit}|$ \\
\midrule
$\sim$200 & within  & 22:10 & 22:10 & $29/32$; $1.16$ \\
$\sim$1590 & beyond & 14:18 & 6:26  & $4/32$; $13.9$ \\
$\sim$3430 & beyond & 6:25\,($+1$ anti) & 3:29 & $10/32$; $13.5$ \\
\bottomrule
\end{tabular}
\end{table}

The within-window row is the positive control: when the retained rows fit inside one window the
compact cache matches native serving on the aggregate follow counts and agrees on $29/32$ first-token
argmax (computed on the ONLINE-donor arm) --- high, but not bit-exact.
The only variable across rows is whether the far row's original distance exceeds $1024$; the sharp
drop in argmax agreement and the order-of-magnitude jump in logit deviation once it does are the
sliding-window mask distortion. This affects only Gemma-4's long-context real-dialog cells, whose
conversations run $21$--$66$k tokens (\S\ref{sec:active}); the short synthetic Gemma-4 runs and all
full-attention (Qwen3) results are unaffected. The gate script is in the release.

\textbf{Serving the real-dialog arm.} Because that distortion is not inert at real-dialog lengths,
the entire Gemma-4 real-dialog arm ($1061$ QA over $20$ conversations, $L$ up to $65.7$k) is served
natively: retained rows stay at their original positions in the full cache, omitted rows are
removed by a key-mask, and the query is appended at the original end position, so the model's own
sliding-window mask applies. The isolated control passes through the identical geometry --- we
write the isolated-encoded carrier KV into its own original slots and restore afterwards --- so the
two arms differ only in the KV content at those slots; both are read by the same first-token
candidate logit. This is also the geometry an eviction-style serving system presents: reference
implementations apply RoPE before compression, so retained keys keep their original rotation, and
SnapKV \citep{snapkv2024} tracks the uncompressed sequence length explicitly so the query stays on
the original timeline.

Serving geometry matters at this scale. Under native serving harvested and isolated land on the
same rate ($.562$ vs.\ $.562$; LoCoMo $-.011$, $95\%$ CI $[-.026,+.004]$), whereas routing the same
items through a compact cache --- retained rows re-packed adjacent to the query --- reports a
deficit ($-.064$; LoCoMo $-.066$, $[-.105,-.031]$). The disagreement is item-level, not a uniform
shift: the two paths return different verdicts on $34\%$ of harvested items and $29\%$ of isolated
ones, and both arms flip on $13\%$. The carrier content is therefore not simply invisible under
either path; what changes is how readable a distant carrier is. Re-packing brings a row thousands of tokens back
adjacent to the query, while native geometry leaves it reachable only through the $8$ global
layers --- consistent with both native arms sitting just above the $.5$ chance of the two-choice
readout. Systems that compact retained rows should therefore expect a different read from systems
that preserve positions.

\section{Decoy design and scoring conventions}
\label{app:honest}

\textbf{The elimination backdoor.} With a binary answer space, a decoy that is always the
complement of gold makes recall and decoy-exclusion indistinguishable: a model that merely excludes
the wrong option scores as though it had recalled the right one, which can carry an effect to
ceiling. All headline audits therefore sample decoys independently, and the binary case is
additionally re-audited in a decoy-stratified manifest. With $r$ estimated from the
$\text{decoy}{=}\text{gold}$ cell ($106/132$; the $123/124$ cell serves as a fit check), a
single-parameter accounting attributes $r\approx.80$ of successes to recall (\S\ref{sec:envelope}).

\textbf{Unscorable readouts.} Transition tables report \textsf{other} outcomes rather than
conditioning them away. In the factorial cells, $3/512$ Qwen3 core-cell readouts were
non-\{ONLINE,OFFLINE\} (dropping $2/128$ items) and $20/768$ on Gemma-4 (dropping $10/192$);
Appendix~\ref{app:tables} re-runs every effect with those readouts counted as failures.

\textbf{Legacy divergence.} On the 2024 checkpoints the passive$\to$\arm{nld} ordering inverts
between the free-generation and extended-readout runs, and the two-model criterion of
\S\ref{sec:active} does not hold on Gemma-2 (Appendix~\ref{app:legacy}).

\section{Updating the served state: current versus history}
\label{app:update}

The update side of the contract is workable and cheap --- provided corrections are served
as \emph{new events} rather than treated as cache edits. We claim no priority on correction or cache
editing itself --- append-erratum and source-influence steering are studied
\citep{mtn2026,kveraser2026}, and explicit serving-side edit/remove/replace directives by
\citet{leyline2026}; we include it to \emph{complete} the write/serve/update
contract and to isolate three update-management effects. For full-attention layers, appending a $p$-token correction
patch repairs the \emph{current} state at prefill cost $O(pL{+}p^2)$ vs.\ $O(L^2)$ for
recomputing an $L$-token prefix: the patch cost stays empirically near-flat across the measured $0.6$--$9.2$k range
($77$\,ms at $L{=}9.2$k), while full recomputation grows to $1033$\,ms at that length. But a served
patch also (i) \emph{hijacks history} --- it pulls historical queries (``the value at step
$t$?'') toward the new value, so corrections must be exposed as \emph{query-scoped versions}
rather than global edits; (ii) \emph{degrades accuracy even when semantically inert} --- even an equal-length
dummy sentence costs $-5.8\pp$ (CI $[-8.6,-2.9]$), while rephrasing a no-op as an explicit
confirmation flips it to a gain; and (iii) is \emph{template-fragile} in natural language
($1.00\to.55$ on current-value readout) but robust as a structured
\texttt{\textless PATCH\textgreater} ($.98$--$1.00$; note this update-side ordering is the
reverse of the active-materialization ordering in \S\ref{sec:active}). Together with
\S\ref{sec:active}: \textbf{write} --- materialize compact derived state into carrier events;
\textbf{serve} --- select the materialized state by query; \textbf{update} --- append new
versions and route historical queries to old ones.

\section{Position-controlled presence \texorpdfstring{$\times$}{x} root-donor replication (X11)}
\label{app:poscontrol}

\S\ref{sec:compose} reports that the root-donor contrast is larger with the downstream
mention absent ($+.398$; $52{:}1$ in Table~\ref{tab:donor}) than co-served ($+.083$). Those two protocols differ in
serve-set membership \emph{and} in the appended query's absolute position: the query is placed
at $\max(\text{served position}){+}1$, the root sits at event slot $13$ and the mention at slot
$18$, so dropping the mention also moves the query five slots closer to the root. X11 separates
the two.

\textbf{Design.} Same 24-event log, Qwen3-8B, $n{=}128$ items ($4$ shards $\times 32$, seeds
$8200$--$8203$). Three serve conditions per root donor: \textbf{men} $=\{$decoy, root, mention$\}$
where the co-served row names the queried register; \textbf{fil} $=\{$decoy, root, filler$\}$
where the \emph{same template} names an unrelated register; \textbf{abs} $=\{$decoy, root$\}$,
the original mention-absent cell. The referent is swapped so that the event tokenizes to
\emph{exactly} the same length (searched over candidate names; $116/128$ items matched and only
these are analysed), so \textbf{men} and \textbf{fil} place the query at an identical absolute
position --- verified per item, $0$ mismatches. Because the mention slot ($18$) follows the root
slot ($13$), causal attention makes the decoy and root rows bit-identical across the two
encodings, so \textbf{abs} is well defined against both. Readouts that are not
\{ONLINE,OFFLINE\} are counted as non-ONLINE (ITT) so all $116$ items are used.

\textbf{Result.} Root-donor effect on $P(\text{ONLINE})$: \textbf{men} $+.086$ ($95\%$ CI
$[.013,.159]$), \textbf{fil} $+.138$ $([.060,.216]$), \textbf{abs} $+.336$ $([.250,.423]$) --- an
independent run whose intervals contain both \S\ref{sec:compose} values (co-served $+.083$,
mention-absent $+.398$), so the estimates are directionally consistent and statistically compatible with the earlier cells. The paired decomposition on the same items:

\begin{center}
\small
\begin{tabular}{lcc}
\toprule
contrast & estimate & $95\%$ CI \\
\midrule
mention effect, position-controlled (men$-$fil) & $-.052$ & $[-.147,+.044]$ ($p{=}.29$) \\
filler-plus-displacement (fil$-$abs)            & $-.198$ & $[-.291,-.105]$ ($p{<}10^{-4}$) \\
original confounded contrast (men$-$abs)        & $-.250$ & $[-.360,-.140]$ ($p{<}10^{-4}$) \\
\bottomrule
\end{tabular}
\end{center}

The confounded contrast reproduces, and roughly four-fifths of it is carried by the
filler-plus-displacement term: serving the tested donor-invariant filler at the mention slot already
costs $-.198$. The mention-specific component is not distinguishable from zero, and the interval is
too wide to certify equivalence at $\pm.075$ (TOST $p{=}.32$), so we report it as no detected effect
rather than proven absence. The root-donor signal is therefore a non-mention-specific serving
effect, carried by the added row and the query displacement it induces.

\section{Addressability protocol, the position control, and the donor-pair floor}
\label{app:address}

\textbf{Protocol.} Trajectories are replayed from the \S\ref{sec:compose} generator with the same
RNG stream, so item $i$ here is item $i$ there (verified $40/40$ identical; an injected wrong-seed
control flags $40/40$, so the check is not vacuous). All $24$ events are retrieval candidates. Each
arm is scored by donor \emph{separation}: the pair counts only if the ONLINE-rendered member answers
ONLINE \emph{and} the OFFLINE-rendered member answers OFFLINE.

\textbf{The decoy is pinned in every arm.} It is the only served event containing a state word, so a
retriever set without it has no answer vocabulary: free generation then returns nothing parseable on
$60\%$ of items, and a first-token candidate probe instead measures decoy echo. The two readouts
fail on the same asymmetry in opposite directions, so neither can compare serve sets until the
scaffold is equalised; the retriever spends its remaining $k{-}1$ rows on evidence.

\textbf{The lexical arm is a null instrument on this template.} Every relevant event ties at two
tokens of query overlap ($P(\text{identical overlap}){=}1.000$), so the ordering is entirely the
recency tiebreak and every retrieval statistic is constant across items (effective $n{=}1$). Read it
as ``no instrument'', not ``no effect''. Separately, indexing against the harness's \emph{full}
question moves the source to rank $0$ --- not through the word \emph{mirror} (there is no stemming:
\texttt{mirror} $\neq$ \texttt{mirrors}) but because the answer-format tokens \emph{ONLINE}/
\emph{OFFLINE} match the source's own value string. Any retrieval evaluation indexed against that
question is measuring the format, not the memory.

\textbf{Role versus position.} The main-text contrast serves the root (slot $13$) against the
reference edge (slot $18$) with the source at $8$--$11$, so role covaries with position and with
distance to the source --- the confound that App.~\ref{app:poscontrol} found to carry most of a
comparable effect. The roles cannot be exchanged: write-time propagation requires the chain in slot
order (``$M_1$ mirrors $S$'' must be encoded before ``$M_2$ mirrors $M_1$''), so in any working
chain the root is necessarily nearer the source. Position can only be \emph{compressed}. We
therefore cross two layouts --- \texttt{spread} $(13,18,23)$ and \texttt{packed} $(20,21,22)$, where
root and edge are \emph{adjacent} and both sit $9$--$12$ slots from the source --- with two seeds,
$n{=}64$ per cell under identical code.

\begin{table}[h]
\centering
\caption{\textbf{The root/edge gap does not depend on the position difference.} Donor separation,
free-generation readout, matched budget. Paired exact McNemar on discordant pairs.}
\label{tab:poslayout}
\small
\begin{tabular}{llccccc}
\toprule
layout & seed & root & edge & discordant & $p$ \\
\midrule
\texttt{spread} & 3900 & $.453$ & $.094$ & $27{:}4$ & $3.4{\times}10^{-5}$ \\
\texttt{spread} & 7717 & $.344$ & $.188$ & $18{:}8$ & $.076$ \\
\texttt{packed} & 3900 & $.328$ & $.125$ & $17{:}4$ & $.0072$ \\
\texttt{packed} & 7717 & $.328$ & $.094$ & $20{:}5$ & $.0041$ \\
\midrule
\texttt{spread} & pooled & $\mathbf{.398}$ & $\mathbf{.141}$ & $45{:}12$ & $\mathbf{1.3{\times}10^{-5}}$ \\
\texttt{packed} & pooled & $\mathbf{.328}$ & $\mathbf{.109}$ & $37{:}9$ & $\mathbf{4.1{\times}10^{-5}}$ \\
\bottomrule
\end{tabular}
\end{table}

Compressing the root-to-edge distance from five slots to one changes the gap by $+.039$
(s.e.\ $.079$, $z{=}0.49$): no detectable interaction, and the gap survives at
$p{=}4.1{\times}10^{-5}$ with the two rows adjacent. Moving the root seven slots further from the
source costs $-.070$ $[-.188,+.048]$, also not resolvable. The seed band is comparable to the
effect --- root separation moves $.453\to.344$ across seeds, and one cell is individually
non-significant --- so we report pooled figures throughout and treat any single $n{=}64$ cell as a
noisier read than it appears.

\textbf{Why reference arms sit below $.25$.} Stated in \S\ref{sec:address}: a reader with no
donor-dependent information echoes the served decoy rather than guessing independently, which floors
separation at $0$ rather than $.25$. Sub-floor cells are therefore the expected signature of rows
carrying nothing donor-distinguishing, not instrument failure.

\textbf{Reference arms and the acceptance test.} The retriever arms compare sparse serve sets
against each other, which cannot price what sparse serving \emph{concedes}. A no-eviction arm
supplies that denominator and the text arms are the null check; all four are reported in
Table~\ref{tab:refarms}. The ceiling there is $.922$, not $1$, because ten of $128$ items fail with
everything served; those failures are not nested inside the sparse arm's (\texttt{oracle} succeeds
on six of the ten), so they are readout noise rather than intrinsically hard items.

The $\texttt{txt}$ arm carries no information about the mechanism and we flag it as such. With the
source omitted, the two donor members' served \emph{text} is byte-identical, so under greedy
decoding the arm is deterministically zero --- this is precisely the operative null stated in
App.~\ref{app:protocols}, not an empirical baseline, and no $p$-value against it would be
meaningful. We include it because that null underwrites every donor claim in the paper and is worth
confirming end to end on this protocol: $128/128$ items produced byte-identical generations across
donors, against $33/128$ for the KV arm on the same served events and $0/128$ for
$\texttt{replay}$, whose served text does differ. A genuine \emph{text} baseline is a different
experiment --- retrieval over a pool that still contains the source, which often simply returns it
(the lexical multi-hop arm, $p_{\text{src}}{=}1.00$) --- and App.~\ref{app:tables} already reports
the source-present case, where clean text re-encoding matches the KV path ($384/384$ vs $377/384$).
The two results delimit the mechanism from both sides: with the source served, text is as good; with
it omitted, text is structurally incapable and the rows are the only route.

Adding these arms is required to leave the pre-existing ones untouched, since they are computed
independently per item from an unchanged RNG stream. That was the acceptance test: $24$ stored
values across the six original arms were compared and $0$ differed. The regression checker was
fault-injected first (three perturbations, including the NaN cell, all caught), so the null result
is informative. Re-parsing the new arms with the independent parser gave $1536$ readouts and $0$
disagreements.

\textbf{Instrument corrections and null checks.} Three issues were found and fixed, two of which
reversed a written conclusion. (i) A candidate-logit readout put the reference arm below its own
floor; the cause was not that the trace instruction pushed the answer out of the first token
(candidate mass at that position is median $1.0000$) but that the first token echoes the served
decoy, which the oracle set contains and the retriever sets did not --- fixed by the anchor above,
not by changing readout. (ii) A floor guard fired on $n{=}4$ noise and now requires $n\ge32$.
(iii) Time-to-first-token had been wall-clocked around a call that also decoded eight tokens against
the text arm's one; using the returned TTFT reversed the sign of the comparison. Two checks came
back clean and are recorded as such: re-running \texttt{spread}/3900 under the modified code
reproduced every stored figure exactly, and re-deriving every separation from stored generations
with an independently written parser (word-boundary regex, last match, fault-injected first) gave
$3072$ readouts with $0$ disagreements.

\section{Real-dialog follow-up: retrieval, retriever-selected rows, and the repaired control}
\label{app:realdialog}

Everything in this appendix runs on Qwen3-8B over LoCoMo-10 \citep{locomo2024} --- ten
conversations, $419$--$689$ turns each, $1{,}986$ QA --- with REALTALK \citep{realtalk2025} as the
second corpus for the repaired control only. Inference is wild-cluster bootstrap-$t$ with the
conversation as the cluster (App.~\ref{app:protocols}); at $G{=}10$ all $1024$ sign vectors are
enumerated, so the smallest attainable two-sided $p$ is $.002$. Every run is greedy and
deterministic: two independent launches of the $k{=}5$ readout agreed on all $4{,}193$ recorded
numbers to the last digit, and at $k{=}1$ the joint and isolated encodings are the same
computation and differ by exactly $0$ on every item, which is used as an assertion throughout.

\textbf{Retrieval ranking (Table~\ref{tab:e1}).} \emph{V} turns state the answer under a
SQuAD-normalised word-boundary match; \emph{M} turns share at least two content words with the
question and state no answer (strict labelling, code only). Timestamp and speaker prefixes are
stripped before matching --- a date answer otherwise matches the session timestamp of every turn,
and naming a speaker otherwise admits all of that speaker's turns as \emph{M} (mean $|M|$ inflated
eightfold). An LLM paraphrase channel ($66{,}742$ turn judgments, $4{,}157$ calls, $0$ failures;
the judge sees question, gold and turn text only) both adds and removes rows relative to the strict
labels (mean $|V|$ $1.72\to1.64$, $|M|$ $12.54\to11.60$), so it is a second calibration and is
reported separately, not as an upper bound; under it the official evidence turn is labelled
\emph{V} on $1{,}162/1{,}531$ questions, the shortfall being paraphrase. Three gates precede every
reported number: per-item rank variance must be non-zero (it is four digits here, against a
literal constant for the lexical arm on the synthetic template); a planted row restating question
and gold must rank first; and appending the gold to the query must move the rank of the best
\emph{V} turn (BM25 $47.6\to1.4$, overlap $56.3\to7.2$, dense $38.0\to13.7$, fusion $27.5\to2.4$,
reranker $23.2\to2.0$). The gates earned their keep twice: a dense arm that returned NaN
similarities on one backend produced a plausible-looking ``weak retriever'' ($\Delta{=}{+}.058$)
and was caught only by the planted row ranking first on $0/200$ items; a shared embedding cache
keyed on conversation but not on text rendering reversed the dense arm's sign ($-.221\to+.071$)
while every gate passed, and was caught by the two lexical arms reproducing to the last digit
while only the three encoder-backed arms moved. The reported numbers were re-derived on a second
machine, interpreter and private cache and agree exactly. The reranker's depth ($50$) is a free
parameter of the deployed stack and is reported with it. The directed graph arm (Graphiti $0.11.3$
over Kuzu, one episode per turn, nine of the ten conversations ingested with a repair pass
restoring rate-limited turns; the tenth hung during ingestion and was dropped for reasons unrelated
to its content) gives $\Delta{=}{-}.244$ overall and $-.041$ $[-.085,+.053]$ on multi-hop, but
ranks the planted row first on only $114/180$ items (every ingested conversation failing), while a
value-bearing turn ranks first on $31.4\%$ of items and an unrelated turn on $41\%$; the gate is
not prejudiced against planted episodes (they rank first \emph{more} often than real value-bearing
turns), so the arm is reported as failing its instrument and its rates are not read. AMA-Bench was
considered as a second corpus and rejected because no turn states a gold answer verbatim
($0/720$), which leaves the estimand undefined there.

\textbf{The repaired positive control (Table~\ref{tab:e3}).} The items are the $62$ QA of
Table~\ref{tab:x8ctrl} (the same enumeration rule, $20$ conversations), extended to $111$ by
lifting the per-conversation injection cap from $4$ to $12$. The published harvest code is
unchanged; the repaired control is a separate module whose frozen pass (\texttt{frz}: frozen
negatives, evidence-only counterfactual, only the true carrier) reproduces the published $50/62$,
$51/62$, $-.016$ exactly under the forced-choice readout. Four repairs, applied to every arm:
\textbf{R1} per-item hard negatives written from question and gold alone (eight candidates each,
filtered against substrings of gold and against values that occur in the conversation), one
chosen among candidates and option orders by the no-context arm's probe as nearest $P{=}.5$ ---
the selector sees no treated arm; \textbf{R2} forced-choice readout from the next-token
distribution after ``Answer:'' (mass on \{A, B\}); \textbf{R3} whole-conversation counterfactual
with an asserted gold residue of $0$ (word-boundary); \textbf{R4} a placebo carrier pointing at
another question in the same conversation. The estimand is
$(q_{\text{orig}}{-}q_{\text{cf}})-(p_{\text{orig}}{-}p_{\text{cf}})$: the true carrier's
original-minus-counterfactual world contrast net of the placebo's, both carriers byte-identical
across worlds. Gates: G1 the no-context arm at chance ($.441$ gold, mean $P{=}.501$, s.d.\ $.114$
--- not a constant, and a probe over the whole candidate pool spreads $P(\text{gold})$ over
$[.06,1.00]$ with absurd negatives rejected at $.91$, so the $.5$ is found, not blind); G2 scorable
$1.000$ (the free-generation readout scored $.045$ of no-context items); G3 residue $0$ (a first
count of $50$ was subword coincidence --- \emph{Yes}/\emph{yesterday}, \emph{Rob}/\emph{problems}
--- and is $0$ under the word-boundary rule the replacement itself uses); I1 the gold-evidence arm
at $.991$, above the $.85$--$.96$ band calibrated on the old readout, reported as is. The
ablations attribute the repairs: R1 alone moves the no-context prior from $.727$ to $.497$; R3
alone doubles the world contrast and restores its asymmetry ($+.013$, $1{:}1$ $\to$ $+.031$,
$3{:}0$ on the same $62$ items; the $36$ items whose counterfactual R3 changes carry
$+.045$ against $+.012$ on the $26$ it does not); R2 moves no number. With ten items affected by an
inherited non-word-boundary replacement removed, the extended estimate is $+.014$ ($p{=}.027$,
$[+.002,+.025]$).

\begin{table}[h]
\centering
\caption{Repairing the injected-carrier control (Qwen3-8B; the $62$ items of
Table~\ref{tab:x8ctrl} and the pre-registered extension to $111$). ``world contrast'' is the true
carrier harvested from the original minus the counterfactual prefill; ``primary'' subtracts the
placebo carrier's world contrast item by item. Intervals are conversation-clustered percentile
bootstrap except the last row's wild-cluster bootstrap-$t$ ($n{=}110$; one item lacks a placebo
readout; $p{=}.040$). The isolated true carrier scores $.823$ under the frozen negatives and $.581$
under the repaired ones. The frozen pass reproduces the published Qwen3 row of
Table~\ref{tab:x8ctrl} exactly.}
\label{tab:e3}
\scriptsize
\setlength{\tabcolsep}{3pt}
\begin{tabular}{llccccc}
\toprule
pass & repairs & $n$ & no-context gold / $P$ & follow:anti & world contrast & primary \\
\midrule
\texttt{frz}  & R2             & 62  & $.742$ / $.727$ & $0{:}0$ & $+.007$ $[-.003,+.018]$ & --- \\
\texttt{cfev} & R1, R2, R4     & 62  & $.452$ / $.497$ & $1{:}1$ & $+.013$ $[+.003,+.025]$ & $+.011$ $[-.002,+.024]$ \\
\texttt{rep}  & R1--R4         & 62  & $.452$ / $.497$ & $3{:}0$ & $+.031$ $[+.006,+.069]$ & $+.028$ $[-.000,+.068]$ \\
\texttt{rep}  & R1--R4, ext.   & 111 & $.441$ / $.501$ & $4{:}0$ & $+.020$ $[+.008,+.034]$ & $\mathbf{+.018}$ $[+.001,+.036]$ \\
\bottomrule
\end{tabular}
\end{table}

\textbf{Retriever-selected rows (Tables~\ref{tab:e2}--\ref{tab:e2k}).} Serve sets come from the
reranked fusion of Table~\ref{tab:e1} at $k{=}5$ over a candidate pool from which every \emph{V}
turn (strict $\cup$ paraphrase channel) has been removed; zero-\emph{V} service is re-verified
after the token cap, and the hot window is $0$ so nothing enters unselected. The $n{=}1269$ items
fall $[127,53,130,168,147,101,112,168,133,130]$ across the ten conversations. Arms serve
byte-identical text at identical original positions: \texttt{none} (header and question only),
\texttt{text} (rows visible in the prompt), \texttt{iso} (each row re-encoded alone at its
position), \texttt{self} (the served rows re-encoded together), \texttt{harv} (rows gathered from
the full-conversation prefill), and the placebo pair \texttt{p\_iso}/\texttt{p\_harv} (rows
retrieved for another question in the same conversation). Primary readout: $24$-token greedy
generation judged by Qwen3-8B; every answer was re-judged by a second model family, whose
disagreements are $86.8\%$ one-directional (stricter, not different), so its arm ordering is
informative. Because the judge bit ties on $87\%$ of items (both arms wrong on $86\%$ of ties ---
the answer is not in the served text by construction), the secondary readout is the forced-choice
log-odds $\log P(\text{gold}){-}\log P(\text{neg})$ against the R1 hard negative, available on
$599$ items; it is a sensitivity instrument and is never reported alone, since it and accuracy
disagree in sign on \texttt{harv}$-$\texttt{iso} ($+1.00$ log-odds at $k{=}5$ against $-.023$
accuracy averaged over $k{=}2,3,5$): reuse buys confidence, not correctness, and the only arm
positive on both rulers is
joint re-encoding, at identical token count and positions. The natural-retrieval condition
(nothing withheld) puts $511/1269$ items ($40.3\%$) at zero served \emph{V} turns with serve sets
identical to the forced ones ($2{,}460$ recorded numbers, $0$ mismatches), and $758$ ($59.7\%$)
with the answer in a served row. The deployed operating point (\texttt{bhot}) gathers the served
rows from the full prefill and prefills the hot tail fresh on top of that view, exactly as the
deployed serving code does; at hot window $0$ it is identical to \texttt{harv}, which is asserted.
Under the hot window the placebo loses discriminating power by construction (real and placebo serve
sets share $66\%$ of rows at $16$ turns), which is why the head channel is
\texttt{bhot}$-$\texttt{self}, whose control needs no placebo.

\begin{table}[h]
\centering
\caption{Retriever-selected rows, value withheld ($n{=}1269$, $10$ conversations, judge readout).
Left: per-arm accuracy. Right: pre-registered contrasts (paired, wild-cluster bootstrap-$t$), the
matched cross-family re-judge ($n{=}1210$ items judged by both), the three-cell split of the
primary contrast, and the power of the wild-cluster test on the actual cluster structure ($400$
synthetic experiments; the pre-registered percentile interval rejects $11\%$ of true nulls).}
\label{tab:e2}
\footnotesize
\setlength{\tabcolsep}{3pt}
\begin{tabular}{lc@{\hskip 8pt}lcc}
\toprule
arm & acc. & contrast & estimate $[95\%]$ & $p$ \\
\midrule
\texttt{none}   & $.162$ & \texttt{harv}$-$\texttt{iso} (primary)        & $-.017$ $[-.042,+.010]$ & $.18$ \\
\texttt{text}   & $.184$ & \texttt{p\_harv}$-$\texttt{p\_iso} (placebo)  & $-.019$ $[-.041,+.005]$ & $.11$ \\
\texttt{iso}    & $.191$ & \texttt{text}$-$\texttt{iso}                  & $-.008$ $[-.037,+.020]$ & $.63$ \\
\texttt{harv}   & $.175$ & primary, matched, first judge                 & $-.012$ $[-.041,+.015]$ & $.34$ \\
\texttt{p\_iso} & $.193$ & primary, matched, second family               & $+.006$ $[-.017,+.029]$ & $.56$ \\
\texttt{p\_harv}& $.174$ & cells abst.$\to$gold / wrong$\to$gold / gold$\to$wrong & $0$ / $94$ / $112$ & \\
                &        & power, $\delta{=}0$ / $.02$ / $.05$ / $.10$   & $.052$ / $.242$ / $.858$ / $1.000$ & \\
\bottomrule
\end{tabular}
\end{table}

\begin{table}[h]
\centering
\caption{The two channels of a harvest, log-odds readout ($n{=}599$ items with a hard negative,
$10$ conversations, wild-cluster $p$). \texttt{self}$-$\texttt{iso} is what co-encoding the served
rows is worth; \texttt{harv}$-$\texttt{self} is what seeing the rest of the conversation adds ---
the materialization channel. At $k{=}1$ the two encodings coincide, so the first column is exactly
$0$ by construction. Lower rows: the same channel under a chain-of-thought readout ($n{=}247$,
paired change against no chain-of-thought) and at the deployed operating point that recomputes a
hot window of $0/4/16$ recent turns ($n{=}559$ common items; accuracy in parentheses; placebo
\texttt{p\_bhot}$-$\texttt{p\_self} at $16$ turns $-0.58$, $p{=}.012$).}
\label{tab:e2k}
\footnotesize
\begin{tabular}{lccc}
\toprule
$k$ & \texttt{self}$-$\texttt{iso} & \texttt{harv}$-$\texttt{self} & \texttt{harv}$-$\texttt{iso} \\
\midrule
1 & $0.000$ & $-0.165$ ($p{=}.21$) & $-0.165$ ($p{=}.21$) \\
2 & $+0.154$ ($p{=}.15$) & $-0.042$ ($p{=}.71$) & $+0.112$ ($p{=}.28$) \\
3 & $+0.480$ ($p{=}.002$) & $-0.008$ ($p{=}.96$) & $+0.472$ ($p{=}.016$) \\
5 & $+1.042$ ($p{=}.002$) & $-0.041$ ($p{=}.85$) & $+1.002$ ($p{=}.002$) \\
\midrule
CoT, paired change & $+0.057$ ($p{=}.09$) & $-0.004$ ($p{=}.92$) & $+0.053$ ($p{=}.16$) \\
\texttt{bhot}$-$\texttt{self}, hot $0/4/16$ & \multicolumn{3}{l}{$-0.04$ / $-0.30$ / $-0.30$ ($p{=}.85/.33/.09$); acc.\ $-.043$ / $-.051$ / $-.032$} \\
\bottomrule
\end{tabular}
\end{table}

\textbf{The narrow form and its cost (Tables~\ref{tab:e2nat}--\ref{tab:e2cost}).} With nothing
withheld, the $394$ items that serve an answer-stating turn and have a hard negative give the
narrow-form contrast \texttt{bhot}$-$\texttt{self}. Its encoding ladder separates the two channels
by row type at hot window $0$: for answer-bearing rows the log-odds run $13.55$ (isolated)
$\to17.14$ (co-encoded, $+3.59$) $\to19.01$ (in place, $+1.87$); for the placebo rows served
beside them $4.65\to5.72$ ($+1.07$) $\to5.85$ ($+0.13$); for the non-value rows of the withheld
condition $2.21\to3.30$ ($+1.10$) $\to3.34$ ($+0.04$). The first step helps every row alike; the
second helps only the row whose own computation depended on what it states. The value swap
re-prefills each conversation with the gold replaced throughout by the item's hard negative (only
the $345$ items whose gold appears verbatim in a served row are swappable; residue asserted $0$;
median one rewritten row) and reads the same served rows in both worlds. Cost is a single-device
microbenchmark (A800, bf16, SDPA, $k{=}5$, $100$ items of one conversation of $15{,}365$ tokens;
median of five after warm-up) and is not serving economics.

\begin{table}[h]
\centering
\caption{The narrow form: in-place KV of an answer-stating served row against joint re-encoding of
the same text (\texttt{bhot}$-$\texttt{self}, log-odds; accuracy at ceiling, $.98$--$.99$, for
every arm). Left: by hot-window size with the placebo contrast. Right: the value swap at hot window
$0$ --- the contrast in the original and the counterfactual world, their difference $I$, and the
placebo's $I$ ($n{=}345$; the counterfactual flips the sign of the readout on $.94$--$.95$ of items,
the positive control). Wild-cluster $p$, $10$ conversations.}
\label{tab:e2nat}
\footnotesize
\setlength{\tabcolsep}{5pt}
\begin{tabular}{lcccc}
\toprule
hot window & $n$ & \texttt{bhot}$-$\texttt{self} & clusters $>0$ & placebo \\
\midrule
0  & 394 & $\mathbf{+1.87}$ ($p{=}.002$) & $10/10$ & $+0.13$ ($p{=}.59$) \\
4  & 386 & $+1.58$ ($p{=}.006$)          & $9/10$  & $-0.02$ ($p{=}.94$) \\
16 & 374 & $+0.67$ ($p{=}.12$)           & $7/10$  & $-0.28$ ($p{=}.27$) \\
0, no unserved \emph{V} turn & 204 & $+1.75$ ($p{=}.004$) & & \\
\bottomrule
\end{tabular}

\vspace{5pt}
\begin{tabular}{lcccc}
\toprule
swap ($n{=}345$) & original & counterfactual & $I$ ($p{=}.002$) & placebo $I$ \\
\midrule
\texttt{bhot}$-$\texttt{self} & $+1.83$ ($p{=}.002$) & $-1.14$ ($p{=}.008$) & $\mathbf{+2.97}$ $[+2.03,+3.94]$ & $+0.02$ ($p{=}.44$) \\
\bottomrule
\end{tabular}
\end{table}

\begin{table}[h]
\centering
\caption{Latency of the serving arms at $k{=}5$ (median ms). Cells are build / TTFT: build $=$
assembling the served cache, TTFT $=$ build plus question prefill and first token; \texttt{tx}
has no build step. Re-prefilling the raw text is fastest at this span; the one-time
full-conversation prefill that \texttt{harv}/\texttt{bhot} amortise costs $3{,}607$\,ms.}
\label{tab:e2cost}
\footnotesize
\begin{tabular}{lcccccc}
\toprule
hot & tokens & \texttt{tx} & \texttt{iso} & \texttt{self} & \texttt{harv} & \texttt{bhot} \\
\midrule
0  & 170 & $70$  & $191$ / $264$ & $45$ / $118$  & $8$ / $81$  & $8$ / $81$ \\
4  & 270 & $78$  & $230$ / $302$ & $55$ / $127$  & $10$ / $84$ & $47$ / $119$ \\
16 & 762 & $130$ & $305$ / $378$ & $129$ / $205$ & $22$ / $96$ & $77$ / $149$ \\
\bottomrule
\end{tabular}
\end{table}

\textbf{Representation-level probe.} On the $111$ control items ($17$ conversations) we prefill
both worlds and measure, per turn, the relative Frobenius drift of its KV
$\|\kv_{\text{orig}}{-}\kv_{\text{cf}}\|/\|\kv_{\text{orig}}\|$ over all layers, keys and values.
A position guard keeps only turns occupying an identical position range in both worlds (the
replacement changes token counts, shifting every later turn; without the guard the apparent
downstream drift is $.140$, two-thirds of it rotation, and the pre-change floor cannot detect it
because it lies upstream of the shift). The floor holds exactly: $19{,}834$ pre-change turns at
$0$. Downstream, $16{,}407$ unchanged turns drift by $.066$ on average (median $.054$), none by
zero, with a local component that decays over the first hundred turns ($.166$ at distance $\le1$)
onto a plateau of $.064$ that does not. Selectivity is absent and inverted: turns sharing $0/1/2$
content words with the question drift $.059/.055/.053$, a ratio of $.917$ for $\ge2$ against $<2$
($.905$ restricted to turns more than $50$ beyond the change). Splitting each turn's drift into
its final newline token and its content tokens, the delimiter drifts $.015$ against $.066$ (its KV
norm is $.83$ of a content token's, so this is not a sink artefact); after residualising on turn
length and distance, delimiter-minus-content drift is $+.148$ ($p{=}.030$) for turns sharing words
with the real question and $+.158$ ($p{=}.027$) for turns sharing words with a placebo question
from the same conversation, so the apparent selectivity is a property of sharing words with any
question. The counterfactual injected carrier of Table~\ref{tab:e3} is the contrast case: a row
written to point at the answer does carry the world difference in its own KV.

\section{Representation-level probes: what the rows hold versus what the model reads}
\label{app:repprobe}

These probes adjudicate the output-interface verdicts of \S\ref{sec:write} and
\S\ref{sec:envelope} at the representation level. A probe is PCA (at most $128$ dimensions)
followed by $L_2$-regularized logistic regression, with four-fold cross-validation grouped by item
(by register-name pair for the payload test). Features are taken every fourth layer, as the mean
and last-token vectors of a row, and each cell reports its best layer. Probes read the
\emph{served} value vectors, which carry no rotary position, and, where stated, the hidden state at
the answer position. Every control also takes its best layer, so a control's rate is the floor of
the instrument rather than chance.

\textbf{Construction bank: writing versus reading (\S\ref{sec:write}).} Design A reuses the bank's
single-source items ($96$ per construction). Design B adds a binding requirement: two sources hold
opposite states and the carrier names one of them at random, so the target is the named source's
state --- an exclusive-or of the name and the first source's state ($256$ items per construction,
$128$ on Qwen3-32B). The row just before the carrier saw both sources but not the name, so it is at
chance by construction, and a linear probe can recover the target from the carrier row only if that
row computed the named source's state. Generation follows the bank's trace protocol on every A item
and on the first $64$ B items; a forced choice between the two state words is scored on every item
(threshold-free AUC). The decoy row precedes every source and gives the causal floor.

\begin{table}[h]
\centering
\caption{\textbf{The carrier row holds the state under every construction.} Ranges are over the
sixteen constructions. Generated: the model's own answer with the source dropped (trace protocol).
$\rho$: Spearman correlation of that rate with Table~\ref{tab:bank}. FC AUC: forced choice between
the two state words. V: probe accuracy on the served value vectors of the carrier row, the row just
before it, and the decoy row. Hidden: probe accuracy on the answer position's hidden state when
only \{decoy, carrier\} are served.}
\label{tab:repprobe}
\footnotesize
\setlength{\tabcolsep}{3.5pt}
\begin{tabular}{llccccccc}
\toprule
Model & Design & Generated & $\rho$ & FC AUC & V carrier & V preceding & V decoy & Hidden \\
\midrule
Qwen3-8B       & A & $.47$--$.93$ & $.90$ & $.57$--$1.00$ & $1.00$        & $.69$--$.86$ & $.51$--$.59$ & $.97$--$1.00$ \\
Qwen3-8B       & B & $.42$--$.75$ & $.80$ & $.78$--$1.00$ & $.99$--$1.00$ & $.48$--$.59$ & $.47$--$.49$ & $.98$--$1.00$ \\
Ministral-3-8B & A & $.48$--$.94$ & $.89$ & $.71$--$1.00$ & $1.00$        & $.56$--$.74$ & $.51$--$.66$ & $1.00$ \\
Ministral-3-8B & B & $.52$--$.98$ & $.81$ & $.98$--$1.00$ & $1.00$        & $.48$--$.60$ & $.46$--$.54$ & $1.00$ \\
Gemma-4-12B    & A & $.42$--$.66$ & $.41$ & $.60$--$1.00$ & $.99$--$1.00$ & $.62$--$.76$ & $.50$--$.67$ & $.85$--$1.00$ \\
Gemma-4-12B    & B & $.39$--$.61$ & $.47$ & $.63$--$1.00$ & $.99$--$1.00$ & $.50$--$.59$ & $.44$--$.52$ & $.92$--$1.00$ \\
Qwen3-32B      & B & $.50$--$.80$ & ---   & $.97$--$1.00$ & $1.00$        & $.51$--$.63$ & $.52$--$.61$ & $.99$--$1.00$ \\
\bottomrule
\end{tabular}
\end{table}

Design A reproduces the bank's profile on Qwen3 and Ministral-3 ($\rho=.90$ and $.89$); on Gemma-4
generation stays compressed near chance, as in the bank. In A the preceding row also decodes the
value ($.56$--$.86$): a single source's value is present in every row downstream of it, so A shows
only that the value is there, and binding is read from B. In B the carrier row decodes the named
source's state at $.99$--$1.00$ for every construction on every checkpoint while the preceding row
stays at its floor, and the answer position holds the state as well; generation varies with the
construction and does not follow the probe. Forced choice reads out more than free generation (AUC
$\ge.97$ for every construction on Ministral-3 and Qwen3-32B in B), the generation--logit split of
regularity (3). Linear decodability shows that the state is present, not that the model can use it.

\textbf{Payload envelope: storage versus readout (\S\ref{sec:envelope}).} On a matched set of
synthetic payload items ($512$ per cell) we probe the served value vectors in three conditions:
the carrier harvested from a prefill with the source present (\emph{carrier}), an unrelated neutral
row at the carrier's slot in the same prefill (\emph{unrelated}), and the carrier encoded without the
source (\emph{unseen}); the carrier's hidden state is probed alongside. The model's own readout on
the same items uses a lettered menu, with a separate two-option question for whether a three-digit
payload exceeds $500$. This menu and item set are not the protocol of \S\ref{sec:envelope}, and the
levels differ (Qwen3's binary readout is $.59$ here against $.934$ there); the comparison that
matters is within items, between the probe and the model's own answer.

\begin{table}[h]
\centering
\caption{\textbf{Payloads are present in the served rows beyond what the model reads back.}
Binary, four-way and eight-way cells are listed as $2/4/8$; ``hundreds'' is the leading digit of a
three-digit payload; ``$>500$'' is a coarse binary property of the same payload. Readout: the
model's own answer with the source dropped (source present in parentheses).}
\label{tab:payprobe}
\footnotesize
\setlength{\tabcolsep}{3.5pt}
\resizebox{\linewidth}{!}{%
\begin{tabular}{llccccc}
\toprule
Model & Payload & Readout (present) & V carrier & V unrelated & V unseen & Hidden carrier \\
\midrule
Qwen3-8B       & $2/4/8$  & $.59/.24/.13$ ($1.0$)          & $1.00/1.00/1.00$ & $1.00/1.00/1.00$ & $.59/.31/.24$ & $1.00/1.00/1.00$ \\
               & hundreds & $.09$ ($1.0$)                  & $.70$ & $.96$  & $.10$ & $.77$ \\
               & $>500$   & $.47$ ($.96$)                  & $.96$ & $1.00$ & $.55$ & $.99$ \\
Ministral-3-8B & $2/4/8$  & $.43/.23/.12$ ($.73/.99/.98$)  & $1.00/1.00/.99$ & $1.00/1.00/1.00$ & $.73/.45/.27$ & $1.00/1.00/1.00$ \\
               & hundreds & $.12$ ($.96$)                  & $.45$ & $.97$  & $.10$ & $.53$ \\
               & $>500$   & $.59$ ($.93$)                  & $.93$ & $1.00$ & $.55$ & $.93$ \\
Gemma-4-12B    & $2/4/8$  & $.89/.30/.17$ ($1.0$)          & $1.00/1.00/1.00$ & $1.00/1.00/1.00$ & $.72/.40/.22$ & $1.00/1.00/1.00$ \\
               & hundreds & $.09$ ($1.0$)                  & $.76$ & $.98$  & $.13$ & $.94$ \\
               & $>500$   & $.42$ ($1.0$)                  & $.97$ & $1.00$ & $.55$ & $.99$ \\
\bottomrule
\end{tabular}}
\end{table}

Categorical payloads are linearly present in the served carrier values at $\approx\!1.00$, while on
these items the model's own readout recovers only Gemma-4's binary payload well ($.89$) and no
four- or eight-way payload; even the binary property ``exceeds $500$'', decodable at $.93$--$.97$,
is answered near its majority rate ($.42$--$.59$). Lower digits
are weak in the carrier (tens and units $.19$--$.38$) but stronger in the unrelated row
($.59$--$.93$), so the value is spread over the rows downstream of its source rather than packed
into the carrier. The \emph{unseen} control decodes above chance because payload length shifts the
carrier's token position, but stays well below the harvested rows. Limits: linear decodability is
not use; the items are synthetic; and Ministral-3's tokenizer emits a regular-expression warning in
this transformers version, identically across arms.

\section{Meaning versus wording on three checkpoints}
\label{app:wording}

The swap probe of \S\ref{sec:discovery} reports Qwen3 alone. Here each claim gets its own donor-pair
contrast, with the logic of the $99{:}0$ test: within a pair every served token and position is
identical, and only the omitted prefix differs. \emph{Meaning} pairs use donor words from opposite
states (e.g.\ \textsc{running} versus \textsc{halted}); among pairs whose two harvested answers
differ we count follow:anti on the state. \emph{Wording} pairs use two words for the same state
(e.g.\ \textsc{running} versus \textsc{online}); we count follow:anti on the word itself. Arms:
\emph{harvest} (carrier rows from one prefill with the donor prefix, prefix omitted), \emph{full}
(everything served, the ceiling) and \emph{isolated} (carrier encoded with no prefix, giving the
model's default word). Counts are sign tests.

\begin{table}[h]
\centering
\caption{\textbf{Meaning crosses on every checkpoint; wording depends on the checkpoint.}
Follow:anti counts among pairs whose two answers differ.}
\label{tab:wording}
\small
\begin{tabular}{lcccc}
\toprule
Model & Pairs & Meaning (harvest) & Wording (harvest) & Meaning / wording (full) \\
\midrule
Qwen3-8B       & $230$ & $56{:}0$ ($p{=}3{\times}10^{-17}$) & $0{:}0$ & $113{:}0$ / $116{:}0$ \\
Ministral-3-8B & $230$ & $45{:}0$ ($p{=}6{\times}10^{-14}$) & $4{:}0$ ($p{=}.12$) & $118{:}0$ / $112{:}0$ \\
Gemma-4-12B    & $238$ & $38{:}0$ ($p{=}7{\times}10^{-12}$) & $52{:}0$ ($p{=}4{\times}10^{-16}$) & $116{:}0$ / $116{:}0$ \\
\bottomrule
\end{tabular}
\end{table}

On Qwen3 and Ministral-3 the harvested answers to a wording pair mostly collapse onto one generic
state word ($85$ of $116$ and $74$ of $112$ pairs), so the string does not cross; on Gemma-4 only
$26$ of $116$ collapse, and in meaning pairs the donor's own word recurs in $124$ of $244$ answers
(against $24$ of $236$ on Ministral-3). The $9.7\%$ of \S\ref{sec:discovery} is therefore Qwen3's
rate. Which aspects of a carried state reach the readout is, like the construction profile of
\S\ref{sec:write}, a property of the checkpoint.

\section{An agent-trajectory side channel on AMA-Bench}
\label{app:ama}

Does gathered KV carry \emph{information} from unselected upstream turns, or diffuse influence? We
take AMA-Bench \citep{amabench} episodes whose trajectory fits the window ($\le16$k tokens) and let a
lexical router choose the kept spans. Five arms answer each question: \emph{text} (re-prefill the
kept spans), \emph{KV} (prefill the trajectory once and gather the kept spans' KV at their original
positions), \emph{isolated} (kept spans encoded alone at their original positions),
\emph{perturb-3} (three unselected upstream turns replaced, length-matched, by another step's text)
and \emph{perturb-all} (every unselected upstream turn replaced, length-matched, by neutral filler;
the kept spans' text is unchanged). Every answer is judged, so a change can be signed: information
predicts an accuracy cost under perturb-all with flips skewed toward right-to-wrong, influence
predicts no cost and symmetric flips. The run covers $948$ questions in $79$ episodes from four
domains, with Qwen3-8B as reader and judge, and is repeated with Qwen3-32B in both roles;
intervals are episode-clustered bootstrap.

\begin{table}[h]
\centering
\caption{\textbf{Gathered KV carries the influence of unselected turns, not their information.}
Accuracy per arm ($948$ questions, $79$ episodes).}
\label{tab:ama}
\small
\begin{tabular}{lccccc}
\toprule
Reader and judge & Text & KV & Isolated & Perturb-3 & Perturb-all \\
\midrule
Qwen3-8B  & $.156$ & $.165$ & $.105$ & $.162$ & $.165$ \\
Qwen3-32B & $.287$ & $.308$ & $.229$ & $.293$ & $.299$ \\
\bottomrule
\end{tabular}
\end{table}

Perturb-all changes $95\%$ of answers on the 8B reader and $93\%$ on the 32B one, yet the accuracy
difference to KV is $+.000$ $[-.024,+.026]$ and $-.009$ $[-.035,+.017]$, with flips balanced ($57$
right-to-wrong against $57$ wrong-to-right, $p{=}1.00$; $85$ against $76$, $p{=}.53$). Perturb-3
changes $81\%$ and $84\%$ of answers, again without a cost ($42{:}40$, $p{=}.91$; $65{:}51$,
$p{=}.23$). What gathering does buy is that the kept spans see one another: KV over isolated is
$+.059$ $[+.032,+.089]$ and $+.079$ $[+.046,+.113]$, while KV over text is $+.008$
$[-.015,+.033]$ and $+.021$ $[+.000,+.043]$. This matches the content-blind drift of the real-dialog
probe (App.~\ref{app:realdialog}) and the fidelity-not-accuracy reading of \S\ref{sec:active}.

\section{Extended material for the main text}
\label{app:extended}

The main text summarizes Sections~1--11. This appendix gives the summarized passages and the tables they rely on in
full, in the order of the main text.

\subsection{Introduction: terminology and identification}
\label{app:ext-intro}

We call the underlying phenomenon \textbf{semantic materialization}: \emph{the
write-time commitment of an upstream-derived state into a downstream event
representation, without requiring the event tokens to repeat that state verbatim.} We
borrow \emph{materialization} as an analogy, not as a claim of database-style stored records or
deterministic query semantics: a downstream event's cached rows can behave like a precomputed
result of upstream computation, recoverable at read time without recomputing it. Although prior work already shows that prefill can write memoized conclusions into downstream notes
\citep{mtn2026}, a harder serving question remains open: what an
\emph{independently served} span carries once its \emph{source is dropped} and the exact future
query was absent at write time.

Accuracy alone cannot measure semantic materialization, because it conflates three
successes: the served text may itself contain the answer, the model may guess from
priors or eliminate decoys, or the rows may genuinely carry hidden state. Our controlled
comparison throughout is the \textbf{donor pair} described above. Decoy elimination is the
subtlest of the three successes, because with a complement decoy, excluding the wrong answer
and recalling the right one look identical; we therefore sample decoys independently
everywhere.

\subsection{The serving contract: predicates and query placement}
\label{app:ext-setup}

\textbf{Three predicates, used throughout.}

\emph{Materializes.} A served span $e$ materializes $v$ for a readout $q$ when its contextualized
rows $\kv(e)$ induce donor-aligned behavior relative to isolated re-encoding of the same text,
with $\esrc$ omitted and $e$'s visible tokens and positions held fixed.

\emph{Trigger and carrier.} The trigger is the write-time construction that makes an event do
this. A carrier is such an event --- one whose $\kv$ holds upstream-derived state without naming
it in text.

\emph{Conditional landing.} A served span is a conditional landing when it materializes $v$ under
span-level ablation: selecting it alongside fixed donor-invariant scaffolding --- a decoy, and an
unrelated register in the active arms --- suffices. This is what separates roots, which land the
value, from reference edges, which mostly route to it. The scaffolding is held constant across
donors, so it cannot explain donor-following; but ``on its own'' throughout means \emph{selected},
not served in isolation.

\emph{Where the query goes.} Retained rows always keep the original positions they were written
at; only the query's placement is a protocol choice, and the two options are not equivalent. Our
synthetic arms append the query after the \emph{last retained} row, which compacts the gap left by
the omitted source. An eviction-style serving system instead leaves the query on the original
timeline: reference implementations apply RoPE before compression, so retained keys keep their
original rotation, and SnapKV \citep{snapkv2024} tracks the uncompressed sequence length explicitly
so that decoding positions do not collapse onto the compressed cache. Placement is not a free
choice: even on short full-attention trajectories, moving the query changes effect size --- serving
a donor-invariant filler at the mention slot costs $-.198$ on Qwen3 (App.~\ref{app:poscontrol}) ---
and on sliding-window checkpoints it additionally changes which rows the local mask can reach, so
those arms are served at original positions (App.~\ref{app:swgate}).

Serve-set effects --- co-serving one span modulating the readout of another ---
are reported separately (\S\ref{sec:active}). The donor-pair construction and operative null,
transition classes, models, decoy sampling, equivalence bounds, compatibility gates, and the full
statistical conventions are in Appendices~\ref{app:protocols} and~\ref{app:honest}.

\subsection{Discovery}
\label{app:ext-discovery}

Concretely (Fig.~\ref{fig:concept}): one event writes ``register $S$ is \textsc{online}'', and a
later event states only ``$M$ mirrors $S$''. We omit the first, serve the second, and ask for
$M$ --- nothing served names \textsc{online}, yet the model answers \textsc{online}, and switches
to \textsc{offline} when we flip the omitted write. That flip is one \emph{donor pair}; across
many, we count among \emph{donor-sensitive} items how many \emph{follow} the omitted value versus
its opposite (\emph{anti}).\footnote{The transition classes \textsf{follow}/\textsf{anti}/\textsf{const}/\textsf{other}, the exact-null argument under greedy decoding, and the sign test are in App.~\ref{app:protocols}.} Two regularities hold (Table~\ref{tab:donor}):

\begin{itemize}[leftmargin=1.4em,itemsep=2pt,topsep=2pt]
\item \textbf{The answer follows the hidden donor.} On Qwen3-8B the split is
\textbf{99:0} ($p{=}3.2{\times}10^{-30}$): $99$ donor-sensitive items answered with the source we
omitted and none with its opposite, though no served span states which value is correct --- this
cell reads out by free generation, and where a menu readout is used instead both candidates
appear symmetrically in the query.

The ratio counts only pairs whose two donor outputs \emph{differ}; donor-blind \textsf{const}
pairs are excluded, so it reports a direction rather than an overall recovery rate. A $256$-pair
replication gives the prevalence behind that direction: $130$ \textsf{follow}, $0$ \textsf{anti},
$125$ \textsf{const}, $0$ identical abstentions, and $1$ discordant unscorable pair. So $131/256$
pairs are donor-dependent, and among them the direction is $130{:}0$
($p{=}1.5{\times}10^{-39}$).\footnote{Qwen3, seeds $3900$--$3907$, root-only serving cell.}
\item \textbf{Donor-following holds across serving setups, models, and readouts.} The rows vary what
accompanies the root --- \emph{root only} serves it as the sole donor-dependent span, while \emph{mention absent} and
\emph{co-served} omit or include a downstream \emph{mention} (a later event naming the queried
register, still without its value) --- and the root's direction never reverses: no cell in
Table~\ref{tab:donor} shows more than a single \textsf{anti} pair, and the narrowest root cell we
measure anywhere, Qwen3 with the mention co-served, is still $39{:}18$ (Fig.~\ref{fig:compose}A).
The direction survives the readout too: under the fixed-candidate menu that Gemma-4 requires,
Qwen3 still follows the omitted source $88{:}0$.
\end{itemize}

\textbf{Semantic polarity survives; verbatim wording survives only on some checkpoints.} The carrier is lossy
and coarse. In a matched swap probe on Qwen3, a hidden donor phrase transferred verbatim in only 9.7\% of
pairs --- indistinguishable from the 8\% verbatim rate of the isolated re-encoding control --- while
its semantic polarity transferred in 64\%. On Qwen3 what crosses is the state, not the string. A paired donor test on three checkpoints (App.~\ref{app:wording}) keeps the polarity result everywhere (follow:anti $56{:}0$, $45{:}0$, $38{:}0$ on Qwen3, Ministral-3, Gemma-4) but makes the string model-dependent: same-state synonym donors are followed $0{:}0$ on Qwen3 and $4{:}0$ on Ministral-3, yet $52{:}0$ on Gemma-4.

\textbf{An explicit update overrides what was carried over.} We also test the reverse: when a
later served event explicitly resets the value and we ask for the current state, omitting the
original source's rows yields no old-value leakage (0/96 items; 95\% upper $3.9\%$), and the served update
wins every time (96/96). The carry-over does not retain a stale value against a fresh
instruction; what persists downstream is only the part already written there.

\subsection{Trigger}
\label{app:ext-write}

\textbf{(1) Same semantics, different fate.} Semantically equivalent constructions that the
model comprehends perfectly still differ by tens of points in write-through: on Qwen3
\texttt{consist} writes at $.94$ but \texttt{write} at only $.52$, and Ministral-3 splits the same
pair $.95$ versus $.56$. The readout gate is \emph{semantic relation $\times$ surface
construction}, not semantics alone. The wording itself carries this, not the company it keeps:
the profile holds in a bank where all sixteen constructions share one background per item and sit
in a carrier slot padded to a common token budget, so the query lands at the same absolute
position and only the surface form varies (spread $.50$--$.95$, \texttt{consist} $.95$ vs.\
\texttt{write} $.58$, comprehension $.97$; Spearman $\rho{=}.88$ against the reported bank,
$t_{14}{=}7.02$), and it reproduces across two checkpoints whose manifests drew independent
backgrounds ($\rho{=}.78$, $t_{14}{=}4.70$). Gemma-4 contributes no ordering to compare: its
trace-readout rates compress into $.41$--$.67$, which is the closed aperture of regularity (2)
rather than a disagreement about which constructions are read back.

\textbf{(2) No universal construction; the readout \emph{aperture} differs by model.} The
\emph{aperture} is the set of constructions whose state the model reads back on a given checkpoint. Its size swings
sharply --- the class totals are $5/4/7/0$ on Qwen3, $7/2/7/0$ on Ministral-3, and $0/2/14/0$ on
Gemma-4 (W/P/U/X) --- so no construction is class W on all three: even \texttt{mirrors}, class W on
Qwen3 at $.80$, never reaches W on Gemma-4. Even a checkpoint's strongest patterns therefore do
not transfer as such to the next.

\textbf{(3) ``Comprehended but not natively generated'' splits at the logit level.} A
forced-choice output-logit probe (calibrated on the text arm) splits the comprehended-but-weak
class into constructions whose candidate logit is reachable though free generation does not
realize it, and constructions undetected even by that probe. Gemma-4 is the clearest recent case:
no construction is class W under the trace readout, yet its candidate logit is reachable on $8$ of
the $16$ --- its two readouts diverge sharply. Whether the
free-generation interface exposes donor-aligned accessibility is thus a \emph{model $\times$
construction $\times$ protocol} property, not a global one (on Qwen3 the probe tracks
generation).

\subsection{Landing}
\label{app:ext-compose}

\begin{table}[h]
\centering
\caption{\textbf{The answer follows the root donor on both checkpoints.} Source-omitted 2$\times$2 donor factorial (root donor $\times$ downstream-mention donor), answer about the mirror register; per-item paired effects on $P(\text{answer}{=}\text{ONLINE})$, complete-case ($n{=}126$ Qwen3, free-generation readout; $n{=}182$ Gemma-4, menu). Interaction $=$ mention effect at root${=}$ON minus at root${=}$OFF. $^{\dagger}$$90\%$ TOST interval; others $95\%$. \textbf{The two root columns are two different contrasts, not one measurement repeated.} They are the mention-absent \emph{root-only} contrast ($n{=}128$/$187$) and the root main effect with the mention \emph{co-served}, measured under serving protocols that also differ in the appended query's position. Estimator and ITT sensitivity: App.~\ref{app:tables}; full four-cell counts:
Table~\ref{tab:factorial_cells}.}
\label{tab:factorial}
\small
\setlength{\tabcolsep}{3pt}
\begin{tabular}{lcccc}
\toprule
& \multicolumn{2}{c}{Root effect} & & \\
\cmidrule(lr){2-3}
Checkpoint & mention-absent & co-served & Mention effect & Interaction \\
\midrule
Qwen3-8B    & $+.398$ $[.311,.486]$ & $+.083$ $[.014,.153]$ & $-.028$ $[-.073,+.017]^{\dagger}$ & $+.008$ $[-.079,+.095]$ \\
Gemma-4-12B & $+.091$ $[.050,.132]$ & $+.220$ $[.175,.265]$ & $+.159$ $[.121,.197]$             & $+.011$ $[-.070,+.092]$ \\
\bottomrule
\end{tabular}
\end{table}

Where does the materialized value live when several later events attended the source? Roots
carry it; reference edges route to them. The sharpest case chains one link further: after the
root ``$M$ mirrors $S$'', a downstream \emph{edge} $\eedge$ (``$T$ mirrors $M$''). Asked about
$T$ with the source omitted, Gemma-4 answers with the \emph{root}'s donor $41{:}0$ --- the
edge routes the query to the root --- yet the edge selected \emph{without} the root carries almost
nothing ($2{:}0$). Ministral-3 reproduces the routing at $298{:}0$, and its edge does carry a
weaker signal of its own ($33{:}16$), so ``edges route'' is the strong pattern rather than a law
(Fig.~\ref{fig:compose}B). A depth-3 chain localises it the same way on Qwen3: served link by link
only the root answers, and dropping the root collapses the chain from $.707$ to $.465$
(App.~\ref{app:tables}). A source-omitted
$2{\times}2$ factorial that swaps the root's donor and a downstream \emph{mention}'s donor
independently maps the rest (Fig.~\ref{fig:compose}A).

\textbf{The answer follows the root donor on both checkpoints, but the two serving protocols size
it differently.} On Qwen3 the root contrast is much larger with the mention absent than co-served;
on Gemma-4 the ordering inverts (Table~\ref{tab:factorial}). That crossover is not mention-specific
on Qwen3. The two protocols differ not only in whether the mention row is served but in the
appended query's absolute position (\S\ref{sec:setup}), and a position-controlled replication ---
query held at a fixed position, swapping only the co-served row's \emph{referent}, token-length
matched --- reproduces most of the drop with a \emph{donor-invariant filler} at the same slot
(App.~\ref{app:poscontrol}). What lowers the root contrast is therefore chiefly the added row and
the displacement it induces --- the filler alone accounts for $-.198$ of the $-.250$ confounded
contrast --- with no additional referent-specific effect detected ($-.052$, $[-.147,+.044]$),
though that interval is too wide to exclude one. On Gemma-4 the mention's own donor instead drives
the answer --- the same mention effect seen in the write contract (\S\ref{sec:write}) and the
serve-set ablation (\S\ref{sec:active}).

\textbf{Serve set and wording both move the readout at read time.} Adding a value-free mention of
the target drops the root's donor-sensitivity rate (Qwen3 $41\%\to22\%$), a drop that tracks the
serve-set change rather than the referent. Wording, by contrast, acts at a fixed serve set:
valenced rather than neutral phrasing in the same slot drives sensitivity to $3\%$ --- read-time
competition, not write-time erasure. A different effect appears when the source \emph{is} served:
oracle-selecting the dependency-chain rows holds multi-hop accuracy nearly flat across depth, far
above the full-context baseline on far spans, but clean text re-encoding of the same rows performs
comparably, so a four-arm decomposition attributes that gain to routing and denoising rather than
to source-omitted materialization (App.~\ref{app:tables}). Those KV rows are a near-lossless
carrier.

\textbf{The root/edge distinction is about role rather than position.} In the link-by-link chain
above, the root also sits nearer the source than the edges do --- unavoidably, since write-time
propagation requires the chain to be encoded in slot order. Role therefore cannot be swapped with
position, only \emph{compressed}: placing root and edge in \emph{adjacent} slots changes the gap by
$+.039$ ($z{=}0.49$) and leaves it significant ($p{=}4.1{\times}10^{-5}$), while moving the root
seven slots further from the source costs $-.070$ $[-.188,+.048]$. The contrast itself also
reproduces under a second readout --- free generation at a matched serving budget, $.398$ vs $.141$,
$45{:}12$, $p{=}1.3{\times}10^{-5}$ --- so it is not an artifact of the menu protocol
(App.~\ref{app:address}). \S\ref{sec:address} takes it to the case where the rows are chosen by a
retriever rather than by us.

\subsection{Access}
\label{app:ext-envelope}

\textbf{Recall, not decoy-exclusion, drives the binary result.} A complement decoy makes recall and decoy-exclusion
indistinguishable, so the binary case is re-audited in a separate decoy-stratified manifest. Harvested recovery clears both heuristic baselines --- echoing the
served decoy, and answering its complement --- by a wide margin, and a method-of-moments split
attributes about four-fifths of the successes to genuine recall
(App.~\ref{app:honest}).\footnote{$229/256{=}.895$ harvested vs.\ $.516$ echo and $.484$
complement; the recall share is $r{\approx}.80$, $95\%$ CI $[.73,.86]$.}

\textbf{A write-time note moves the derived frontier, not the numeric one.} A note asking the
model to compute the verdict at write time lifts \emph{derived} recovery well above its
no-prompt level ($.542\to.703$), yet leaves \emph{harvest-dropped} numeric payloads unchanged
($0\to0$) --- even though the identical note rescues that same computation when the source is
still present ($.401\to.938$). The write policy is movable, but this intervention did not extend to
source-omitted numeric recovery, and for derived verdicts it carries a full-context cost
(\S\ref{sec:limits}).

\textbf{Query form matters as much as the store.} A first-digit probe ($.20$) or a quantile query
($.62$) reads out well above exact numeric recall, and two compact bindings carry in parallel; the generation/logit split of \S\ref{sec:write} recurs
(Fig.~\ref{fig:envelope}C). Access is itself model- and construction-dependent: several latent
carriers answer a \emph{source}-addressed query but stay weak under a \emph{target}-addressed
one (App.~\ref{app:protocols}). The primitive is thus \emph{not reliably target-addressable}: its preferred address varies by
construction.

\subsection{Capacity is not addressability}
\label{app:ext-address}

\S\ref{sec:envelope} closes on an addressing problem internal to the query. A serving system faces
the external one: it is never handed the rows that carry the value, it has to find them in an index
built over event \emph{text}. Writing end-to-end accuracy as
$A_{\text{e2e}}{=}P(\text{hit})A_{\text{hit}}{+}(1{-}P(\text{hit}))A_{\text{miss}}$, everything above
measures $A_{\text{hit}}$ and sets $P(\text{hit}){=}1$ by construction. Here we remove that
assumption on the donor-paired register trajectories ($n{=}64$ per cell), holding every retriever to
a matched budget $k{=}|\text{oracle set}|$ so no arm buys separation with a wider window. All $24$
events are candidates: \emph{the source is in the pool}, and we measure whether a retriever evicts
it rather than assuming it away. The results below are ordered by how far they travel --- a
structural claim first, then what the channel is worth, then a retrieval demonstration in which only
one of the two stages is a valid instrument, then two evaluation traps that transfer off this
corpus, then the same question put to a full deployed stack on real long-term dialogs. Cost is a
separate axis and is reported last.

\textbf{The gap: what the model separates, a text index cannot.} On the index side, ``$A$ mirrors
$B$'' and ``$B$ mirrors $A$'' are the same bag of words, and a bi-encoder places a root and the edge
pointing at it at near-identical similarity. This is a property of what the index \emph{encodes},
not of our phrasing, and for the lexical family it is provable rather than empirical. Restricted to
the query's terms the two rows carry \emph{identical} term multisets --- both contain
\texttt{register} twice and the queried name once, and the tokens that differ appear in neither
query --- so any scorer that is a function of the term multiset must return the same score. Okapi
BM25 does: on every item the root and the edge score bit-identically (e.g.\ $2.3393331728978253$ for
both), the tie falls to the recency order, and the edge --- being later --- wins on $200/200$ items.
Term weighting, term frequency and length normalisation are all powerless here; no wording makes a
text index represent the direction of a reference edge. On
the model side the two are plainly different. Holding items, budget and scaffold fixed and varying
only which of the two mirror statements is served, root separates donors at $.398$ against $.141$
for the edge ($45{:}12$ discordant, exact McNemar $p{=}1.3{\times}10^{-5}$, $n{=}128$, free
generation). App.~\ref{app:tables} already separates the chain link by link with the source dropped,
at a larger $n$ and a tighter $p$ --- root $.699$ against $.465$ and $.484$ for the two edges, level
with that experiment's $.473$ text floor, $n{=}256$ under a binary menu --- so the contrast is
readout-independent and should be cited there; the present run adds its robustness, not the result.

What is new is the position control. Root and edge differ in slot as well as in role, and
App.~\ref{app:poscontrol} found position carrying most of a comparable effect. The roles cannot be
exchanged: write-time propagation requires the chain in slot order (``$M_1$ mirrors $S$'' must be
encoded before ``$M_2$ mirrors $M_1$''), so in any working chain the root is necessarily nearer the
source, and position can only be \emph{compressed}. Crossing two layouts --- \texttt{spread}
$(13,18,23)$ and \texttt{packed} $(20,21,22)$, where root and edge are \emph{adjacent} and both sit
$9$--$12$ slots from the source --- compressing that distance from five slots to one changes the gap
by $+.039$ (s.e.\ $.079$, $z{=}0.49$) and it survives at $p{=}4.1{\times}10^{-5}$; moving the root
seven slots further from the source costs $-.070$ $[-.188,+.048]$, also not resolvable. The gap is
structural on both sides: carrying and pointing are behaviourally distinct in the model and
indistinguishable in the index.

\textbf{What the channel actually buys.} Retriever arms compared against each other cannot price
what sparse serving \emph{concedes}; that needs arms we pin rather than retrieve
(Table~\ref{tab:refarms}). Serving \emph{all} $24$ events --- evicting nothing, so the source is
present --- separates donors at $.922$, and serving the carrier alone recovers $43\%$ of that. The
ceiling is $.922$ rather than $1$ because ten of $128$ items fail with everything served, and those
failures are not nested inside the sparse arm's (\texttt{oracle} succeeds on six of the ten), so
they are readout noise and sparse arms should be read against $.922$. The one-line reading is that
the same two events are worth $.000$ re-encoded as isolated text and $.398$ served as KV: the
channel is causal, and it is not a free substitute for the source. \S\ref{sec:discovery} should not
be read as one.

\begin{table}[h]
\centering
\caption{\textbf{Reference arms, pooled over seeds ($n{=}128$, layout \texttt{spread}).} Separation
counts a pair only if the ONLINE-rendered member answers ONLINE \emph{and} the OFFLINE-rendered
member answers OFFLINE. ``echo'' is the fraction of items on which both donor members receive an
\emph{identical} answer --- the signature of a reader with no donor-dependent information, which
floors separation at $0$ rather than $.25$.}
\label{tab:refarms}
\small
\begin{tabular}{llccl}
\toprule
arm & served & sep & echo & role \\
\midrule
\texttt{replay} & all 24 events, KV & $.922$ & $.07$ & ceiling: nothing evicted \\
\texttt{txtrep} & all 24 events, text & $.930$ & $.06$ & ceiling as one-shot prefill \\
\texttt{oracle} & $\{$decoy, root$\}$, KV & $.398$ & $.59$ & the served set of \S\ref{sec:compose} \\
\texttt{txt} & $\{$decoy, root$\}$, text & $.000$ & $.98$ & the operative null, not a baseline \\
\bottomrule
\end{tabular}
\end{table}

The $.000$ is the operative null of App.~\ref{app:protocols} rather than a result: with the source
omitted both donor members receive byte-identical text, so greedy decoding makes the arm
deterministically zero, and no $p$-value against it is meaningful. We report it only as a check that
the null holds, verified $128/128$. Cells below $.25$ are likewise expected rather than broken. The
$.25$ floor assumes a reader guessing \emph{independently} on each donor member; a reader with no
donor-dependent information instead echoes the served decoy, which is identical across members, so
both receive the \emph{same} answer and the pair can never separate --- that strategy floors at $0$.
Measured echo is $74\%$ (\texttt{spread}) / $79\%$ (\texttt{packed}). The $.25$ guard is retained for
the oracle arms, all of which clear it ($.281$--$.453$).

\textbf{Two retrieval stages, and what they are not.} We attach the two cheapest stages a memory
system runs --- an unweighted token-overlap index and Qwen3-Embedding-0.6B, the dense stage of
current agent-memory systems --- and state their status before their numbers. These are a floor, not
a representative stack: deployed memory systems fuse BM25 with dense retrieval by reciprocal rank
and add a cross-encoder reranker, and several maintain a graph layer of directed entity relations
alongside the text index \citep{zep2025,amem2025}. \emph{The unweighted lexical arm is a null
instrument on this template}: every relevant event ties at two tokens of query overlap
($P(\text{identical overlap}){=}1.000$), so the ordering is entirely the recency tiebreak and every
retrieval statistic is constant across items (effective $n{=}1$). Read it as ``no instrument'', not
``no effect''; we report it only as the negative control for the BM25 result above, which does not
depend on the tie being unweighted. \emph{The dense arm does measure}, and what it finds is that one
hop works and two do not. Where the query names the carrier's own subject it finds it (mean rank
$0.17$ of $24$, $95\%$ CI $[0.12,0.22]$; $r@1{=}.83$), selecting the oracle set on $52/64$ items and
separating at $.422$ against the oracle's $.453$. One hop further out the carrier falls below an
irrelevant filler (rank $10.89$, CI $[9.89,11.89]$, vs $1.98$) and exact reconstruction of the
four-row oracle set is $.13$ at $k{=}8$; BM25, by contrast, places the multi-hop carrier at rank
$4.0$, above the best filler at $5.0$, so the multi-hop pessimism is specific to the dense stage
rather than to text indexing as such. Structure in the query is what makes the carrier addressable:
rewriting the query to name the mirror relation moves oracle exact@$8$ from $.13$ to $.66$ ($.48$ to
$.94$ at one hop) --- knowledge a deployed system does not have. The zero variance that makes the
lexical arm uninterpretable is a property of that \emph{family}, not of the template: the term
multiset each item presents to its query is the same up to renaming, so the lexical arm returns the
identical rank on all $200$ items (variance $0$, one distinct value), while the dense arm produces
$200$ distinct orderings and $199$ distinct role-to-rank tuples on the same items --- which is why
its rates carry intervals and the lexical arm's do not. The dense rates remain properties of the
synthetic template and are biased in both directions --- the one-hop rank is flattered by a template
that guarantees query and carrier share a register name, the multi-hop rank punished by fillers
nearly interchangeable with the query. Read the direction, not the magnitude. The exact-tie result above is the one retrieval finding here that is structural rather
than a rate.

\textbf{Two evaluation traps.} Neither depends on our template. \emph{A retriever can succeed
without using the mechanism at all.} If the pool still contains the source, selecting it makes the
answer ordinary reading. This is not an edge case: the lexical arm serves the source on $64/64$
multi-hop items and posts the best multi-hop number in the run ($.547$) without ever exercising
materialization, and the dense arm serves it on $44\%$, whose removal drops it from $.391$ to
$.194$. Any end-to-end memory evaluation on a corpus that retains the source can be won this way, so
we report every arm twice, with and without the source-serving items; only the latter is
mechanism-dependent. \emph{Indexing against the full question measures the answer format, not the
memory.} Retrieved against the harness's complete question, the source jumps to rank $0$ --- not
through the word \emph{mirror} (there is no stemming: \texttt{mirror} $\neq$ \texttt{mirrors}) but
because the answer-format tokens \emph{ONLINE}/\emph{OFFLINE} match the source's own value string.
Any retrieval evaluation indexed against a question that carries its answer options is measuring
that string match.

\textbf{The same question on real dialogs.} The template cannot say how often a deployed index
actually withholds the row that carries the value; LoCoMo \citep{locomo2024} can. For each of $555$
questions over ten long-term conversations ($419$--$689$ turns each) we label every turn as
\emph{V} (states the answer), \emph{M} (shares at least two content words with the question but
states no answer) or unrelated, and ask whether an index ranks some \emph{M} turn above every
\emph{V} turn --- the real-corpus form of routing-over-carrying, since the strict root/edge
distinction needs donor pairs that only a synthetic corpus affords. Because $|M|$ is about ten
times $|V|$, the raw rate is near $.76$ under a random ordering, so Table~\ref{tab:e1} reports
$\Delta{=}\pi{-}\text{chance}$ per item: negative means the index finds the value-bearing turn
better than chance, positive that it \emph{prefers} the topical one. The stack is the one
agent-memory systems deploy, added a stage at a time --- unweighted overlap as a negative control,
Okapi BM25, Qwen3-Embedding-0.6B, reciprocal-rank fusion of the two, and
\texttt{bge-reranker-v2-m3} over the fused top-$50$ --- and each arm passes three gates before its
number is read (App.~\ref{app:realdialog}). Every arm beats chance overall and the full stack does
best --- $-.12$ for BM25, $-.22$ for the dense stage, $-.42$ for the reranked fusion (fusion alone
is worse than dense alone). The split by question type is the result.
On single-hop questions the full stack reaches $-.50$; on multi-hop questions no arm is below
chance --- the reranked fusion sits at $-.008$ $[-.114,+.174]$ --- and the term-frequency family
actively prefers the topical turn (BM25 $+.264$ $[+.201,+.377]$ under the strict labelling of
\emph{M}; at chance under an LLM labelling, so this one reading carries its labelling). The miss is
not scarcity: raising the number of value-bearing turns from one to four or more leaves BM25
unmoved ($\pi{=}.634/.638/.655$) while the dense arm improves ($\pi{=}.561/.534/.491$; lower is
better), and
rank fusion re-imports the lexical arm's multi-hop bias (dense $-.024$ to fused $+.128$, interval
excluding zero). Since the same components work well on single-hop questions, the multi-hop
ceiling is a property of text indexing on this corpus, not of tuning. It also makes the
source-withheld condition of \S\ref{sec:active} the natural multi-hop case: on $53$ of $59$
multi-hop questions the value-bearing turn is buried, and $81\%$ of the winning topical turns lie
downstream of it --- but no further downstream than the pool as a whole ($+.006$ over the base
rate) and no more lexically tied to the value's event than a random downstream turn ($-0.64$
shared content words, interval including zero), so the retriever is blind to position and to
reference. A directed graph index (Graphiti over Kuzu) ran on the same items but failed its
planted-row gate ($114/180$ first, every ingested conversation failing), so its rates are not
reported and the structured-index question stays open.

\begin{table}[t]
\centering
\caption{\textbf{On real dialogs a deployed text stack finds the answer-bearing turn on single-hop
questions and not on multi-hop ones.} LoCoMo-10 ($10$ conversations, $419$--$689$ turns each);
$\Delta{=}\pi{-}\text{chance}$, where $\pi$ is the probability that some topical answer-free turn
(\emph{M}) outranks every answer-stating turn (\emph{V}) and chance $=|M|/(|M|{+}|V|)$ per item
($.761$ on average); negative finds the value better than chance, positive prefers the topical
turn. Strict entity-match labelling, $n{=}555$ items with both roles present; conversation-clustered
bootstrap $95\%$ intervals ($B{=}4000$, seed pinned). Arms: unweighted token overlap (negative
control), Okapi BM25, Qwen3-Embedding-0.6B cosine, reciprocal-rank fusion of BM25 and dense
($c{=}60$), and \texttt{bge-reranker-v2-m3} over the fused top-$50$. Every arm passes a per-item
rank-variance gate and a format-leak probe; the last column is the gate that a planted row restating
question and gold must rank first. An LLM labelling of \emph{M} ($n{=}1208$, chance $.802$) gives
the same sign overall (BM25 $-.201$, reranked fusion $-.513$) but puts BM25's multi-hop preference at
chance ($-.012$ $[-.072,+.047]$), so that cell carries its labelling; under it the reranker fails the
planted gate ($188/200$) and its row is not read.}
\label{tab:e1}
\footnotesize
\setlength{\tabcolsep}{3.5pt}
\begin{tabular}{lcccc}
\toprule
arm & $\Delta$, all ($n{=}555$) & single-hop ($n{=}461$) & multi-hop ($n{=}59$) & planted first \\
\midrule
\texttt{overlap} & $-.122$ $[-.175,-.066]$ & $-.186$ $[-.251,-.120]$ & $+.213$ $[+.150,+.281]$ & $200/200$ \\
\texttt{bm25}    & $-.124$ $[-.180,-.064]$ & $-.197$ $[-.268,-.122]$ & $+.264$ $[+.201,+.377]$ & $200/200$ \\
\texttt{dense}   & $-.221$ $[-.276,-.171]$ & $-.253$ $[-.311,-.198]$ & $-.024$ $[-.129,+.154]$ & $195/200$ \\
\texttt{rrf}     & $-.181$ $[-.231,-.124]$ & $-.231$ $[-.293,-.163]$ & $+.128$ $[+.028,+.278]$ & $200/200$ \\
\texttt{rerank}  & $\mathbf{-.424}$ $[-.471,-.372]$ & $\mathbf{-.496}$ $[-.547,-.442]$ & $\mathbf{-.008}$ $[-.114,+.174]$ & $191/200$ \\
\bottomrule
\end{tabular}
\end{table}

\textbf{A separate axis: keeping rows versus keeping a summary.} Even where addressability is
solved, the row has to be worth storing. At large serving budgets, gathered KV reaches
time-to-first-token speedups of $10.5\times$ at $9$k served tokens and $16.5\times$ at $32$k against
re-encoding the same raw span. A deployed alternative would keep a \emph{summary} rather than that
span, and inverting the measured text curve a summary ties the KV arm at $\approx\!12\times$
compression --- a requirement that \emph{saturates} rather than tracking the speedup, because the
gather scales with the served span too ($143$ of the $286$\,ms at $32$k). Pricing out that copy, as
a paged-attention server would, roughly doubles it to $20$--$25\times$. Storage is the other side:
$144.0$\,KB/token of KV against $4.8$ bytes/token of text, a flat $\sim\!30{,}300\times$ that accrues
per unit \emph{time} rather than per query. We report this as a latency microbenchmark on one device
at batch $1$, not as serving economics --- no arm here is scored for accuracy, and the paper's own
text baseline is the more accurate one.

\textbf{What this bounds.} Four things travel: the structural claim (a text index cannot separate
carrying from pointing), the model-side role contrast, which involves no retriever, the two
evaluation traps, and the real-dialog rates of Table~\ref{tab:e1}, which are properties of one
corpus and ten conversation clusters rather than of the template. The synthetic retrieval rates do
not, for the reason given above. The negative stays bounded: the provable half covers the
term-frequency family, which includes BM25; on real dialogs the measured half now covers the four
text stages a deployed stack runs, at rank level --- whether an unfound value costs an answer is the
question \S\ref{sec:active} takes up on the rows such a stack actually selects. A \emph{graph}
index, which stores entity relations with a direction and could in principle represent exactly the
distinction a text index cannot \citep{zep2025}, did not pass its instrument gate here, and no
multi-hop router was tested, so the claim is about text indexes, not about retrieval. The practical
reading is a scope statement rather than a verdict --- materialized rows pay where the source is
genuinely \emph{gone}, evicted or compacted; where it merely has not been retrieved yet, fetching
it is both available and better, and on single-hop questions a deployed stack does exactly that.

\subsection{Programming the primitive}
\label{app:ext-active}

\textbf{A deliberate carrier induces donor-aligned recovery.} In donor-paired synthetic
trajectories we emit the carrier ourselves --- one answer-free event after the source, drawn from
five arms running from a passive mention to an explicit-text upper bound. Swapping the passive
mention for an answer-free NL-compute directive (``determine the verdict now and record it here'')
lifts donor-aligned recovery from $6\%$ to $51\%$ on Qwen3 without ever naming the value
(Fig.~\ref{fig:active}B, Table~\ref{tab:x9}); leakage and echo controls are clean and
\textsf{anti}-transitions stay negligible. Because those arms differ in carrier length, the
serve-set ablation supplies the position-controlled form: with the query at a fixed absolute
position, the shared downstream review row alone yields $.00$ on Qwen3, while adding the carrier
beside it yields $.42$ under the NL directive and $1.00$ under explicit text
(Table~\ref{tab:x10}). The direction is decisive --- \S\ref{sec:compose} showed that merely adding
a row \emph{lowers} donor sensitivity ($41\%\to22\%$), so a rise from zero is not extra-row
interference. The carrier row is what carries.

\textbf{The effective carrier is a per-checkpoint choice, not a universal syntax.} Which
construction writes depends on model and readout --- Qwen3 resolves a non-saturating ordering,
Gemma-4 ceilings on every arm, Ministral-3 is abstention-dominated under free generation --- so no
materialization syntax is universal, and the preregistered two-model criterion did not carry over
(App.~\ref{app:legacy}). The one arm that recovers on every recent model is an explicit textual
record. Hence the recipe: \emph{emit compact materialization events for state needed after its
operands leave the serving set, calibrate the construction per checkpoint, and keep an explicit
textual record as the fallback.}

\textbf{Natural-dialog stress test.} On REALTALK and LoCoMo long-term dialogs
\citep{realtalk2025,locomo2024} we served answer-free late mentions harvested from the
full-history cache against isolated encodings of the same text, with the sliding-window checkpoint
served at its original positions --- the geometry an eviction-style system presents
(App.~\ref{app:swgate}). Harvesting yields no detectable
benefit on either recent model (Table~\ref{tab:x8}); on Gemma-4 it is \emph{equivalent} to isolated
encoding within the $\pm.05$ band (LoCoMo $-.011$, $95\%$ CI $[-.026,+.004]$).\footnote{REALTALK contributes a descriptive point estimate only:
its ten dyads form three participant components, not ten independent units
(\S\ref{sec:limits}).} The Qwen3 null survives
its own control: with YaRN bringing every conversation inside the native window the deficit widens
rather than closing (\S\ref{sec:limits}). The frozen same-path injected-carrier control did not
fire cleanly on either model (Table~\ref{tab:x8ctrl}); repaired, it fires on Qwen3 (next
paragraph), so the null is bounded by effect size rather than by the instrument:
\emph{passive contextual harvesting is not a dependable end-to-end write interface}.

\textbf{The positive control, repaired.} The frozen control could not fire on Qwen3 for reasons
that belong to the instrument, not the mechanism, and each was fixed under pre-registration and
applied to every arm alike (App.~\ref{app:realdialog}). The wrong option had been borrowed from
another question's gold and was rejectable from the question alone --- the no-context arm's
probability on gold averaged $.727$ --- so per-item hard negatives are now written by a model that
sees only the question and its gold, and one is selected, blind to treatment, by a no-context
probe to sit near $.5$. The free-generation readout had left over half of Gemma-4's
counterfactual items unscorable ($33/62$),
so the readout is the next-token distribution after ``Answer:'' (scorable $1.000$; the old readout
scored $.045$ of no-context items). The counterfactual world had replaced the value only in the
official evidence turn, leaving $1{,}210$ unreplaced mentions across $36$ of $62$ items, so the
replacement now spans the whole conversation with an asserted residue of zero. And a placebo
carrier pointing at another question in the same conversation supplies the diffuse-contamination
baseline. Because the injected carrier restates the question, isolated encoding of it alone already
lifts $P(\text{gold})$ from $.50$ to near $1$, so the estimand is the world contrast --- the same
carrier harvested from the original versus the counterfactual prefill --- net of the placebo's.
The frozen pass reproduces the published $50/62$, $51/62$ and $-.016$ bit-for-bit under the new
readout, so what moved is the negatives and the counterfactual, not the scoring. On the published
$62$ items the world contrast is $+.031$ $[+.006,+.069]$ with follow:anti $3{:}0$ (the frozen
counterfactual gave $+.013$ and $1{:}1$); on the pre-registered power extension to $111$ items the
placebo-corrected contrast is $\mathbf{+.018}$ (wild-cluster bootstrap-$t$ over $16$
conversations, $p{=}.040$, $95\%$ CI $[+.001,+.036]$; follow:anti $4{:}0$ against the placebo's
$1{:}2$), replicating on LoCoMo ($+.019$, $n{=}63$) and REALTALK ($+.017$, $n{=}47$) with every
leave-one-conversation-out estimate positive. Its shape belongs with it: the median per-item effect
is $0.000$ and trimming five items from each tail leaves $+.006$ (sign test $p{=}.11$), so a
deliberately pointed
carrier materializes as a sparse event --- most rows carry nothing, a few carry cleanly --- while
the no-context floor stays at chance ($.44$ gold, mean $P{=}.50$) and the gold-evidence arm rises
to $.991$, above the $.85$--$.96$ band calibrated on the old readout. The path transmits, and the
real-dialog nulls that follow are nulls of effect, not of instrument.

\textbf{Rows a retriever selects.} The harvest above chose carriers by our heuristic. We let the
strongest stack of \S\ref{sec:address} (reranked fusion, $k{=}5$) choose them on LoCoMo,
withholding every value-bearing turn from its candidate pool so that each of $n{=}1269$ items over
$10$ conversations serves zero such turns --- verified after budget capping and through an LLM
paraphrase channel --- with the hot window off so nothing enters unselected. Six arms serve
byte-identical text at identical positions and differ only in where the KV was computed: isolated
per row, jointly over the served rows, or harvested in place from the full-conversation prefill,
each with a placebo set retrieved for another question. Harvested minus isolated is $-.017$
(wild-cluster bootstrap-$t$, $p{=}.18$, $95\%$ CI $[-.042,+.010]$); the placebo contrast is $-.019$
and tracks it; abstention-to-gold conversions number $0$; and re-judging all $7{,}614$ answers with
a second model family leaves the null and flips its sign ($-.012$ to $+.006$ on the matched
$1{,}210$ items), so $|\delta|$ is bounded within about $\pm.03$. A power check on the actual
cluster structure rejects $\delta\ge.05$ with power $.86$ and cannot reject $.02$. The withheld
condition is not a construction: with nothing withheld the same stack serves zero value-bearing
turns on $40\%$ of items, with serve sets identical to the forced ones. A log-odds readout, which
resolves differences the judge bit cannot (the judge ties $87\%$ of items), splits the harvest into
two channels: co-encoding the served rows together is worth $+1.04$ log-odds over isolated encoding
at $k{=}5$ and grows with $k$, but seeing the rest of the conversation adds nothing at any $k$
($-.17/-.04/-.01/-.04$ for $k{=}1,2,3,5$, every interval including zero), under a chain-of-thought
readout (paired change $-.004$, $p{=}.92$), and at the deployed operating point that recomputes a
hot window of recent turns ($-.04/-.30/-.30$ log-odds at $0/4/16$ turns, never positive on either
ruler, the placebo moving as much or more). The representation says the same: changing one upstream
fact moves every downstream turn's KV by $6.6\%$ on average, yet turns sharing content words with
the question move \emph{less} (ratio $.92$), and drift on newline delimiters is about a quarter of
that on content tokens, with a placebo question reproducing its apparent selectivity. Natural
downstream
turns absorb a diffuse, content-blind perturbation, not the withheld value: on real dialogs the
trigger side of the contract holds.

\textbf{What a harvested row does carry: its own context.} Withholding the value tests the wide
form of the contract --- a row acquiring state from another row --- and it is zero. The narrow form
is what the contract predicts should survive: a row's KV computed in place carries what \emph{that
row's own} computation depended on. Letting the same stack retrieve freely, $60\%$ of items serve
a turn that states the answer ($n{=}394$ with a hard negative); accuracy is at ceiling for every
arm ($.98$--$.99$), and the in-place KV of the served rows reads the stated value with
$\mathbf{+1.87}$ log-odds more confidence than the same text jointly re-encoded (wild-cluster
$p{=}.002$; median $+2.0$; sign $263{:}125$; positive in all ten conversations), against $+0.13$
for the placebo rows served beside them ($p{=}.59$). It is not evidence gathered from unserved
value turns --- the effect is $+1.75$ when none exist --- and it is evidence-driven rather than a
prior toward gold: swapping the value throughout the conversation and re-prefilling mirrors it,
$+1.83$ in the original world and $-1.14$ in the counterfactual, an interaction of $+2.97$
$[+2.03,+3.94]$ against a placebo interaction of $0.02$ ($n{=}345$). In-place KV is a
higher-fidelity representation of what a row says --- it follows that row whichever value it
states --- which is the sense in which KV exceeds text: fidelity, not accuracy. The recomputed hot
window partly overwrites it ($+1.58$ at $4$ turns, $+0.67$ at $16$), and at this scale the gain is
not latency: at $k{=}5$, re-prefilling the text reaches the first token in $70$\,ms against $81$
for the in-place gather and $118$ for joint re-encoding (one device, batch $1$), so text is faster
by $10$--$40$\,ms across hot-window sizes while joint re-encoding is slower by $8$--$57$; the
conversation's one-time prefill costs $3.6$\,s. Keeping the rows buys fidelity, not time, until
the served span is large (\S\ref{sec:address}). The same reading holds on agent trajectories: on AMA-Bench, replacing every unselected upstream turn with length-matched neutral filler changes $93$--$95\%$ of gathered-KV answers yet moves accuracy by under a point, with right-to-wrong and wrong-to-right flips balanced, for an 8B and a 32B reader alike (App.~\ref{app:ama}).

\textbf{Updating the state costs a patch, not a recomputation.} Corrections are cheap served as
\emph{new events} rather than applied as cache edits: for full-attention layers, appending a
$p$-token patch repairs the current state at $O(pL{+}p^2)$ versus $O(L^2)$ for recomputing an
$L$-token prefix. But a served patch hijacks historical queries unless query-scoped, degrades
accuracy by its mere presence, and is template-fragile (App.~\ref{app:update}). The triad is
therefore \textbf{write} compact derived state into carrier events, \textbf{serve} it by query, and
\textbf{update} by appending versions and routing historical queries to the old ones.

\subsection{Consequences}
\label{app:ext-consequences}

The contract is directional. A row materializes state that its own computation depended on; it
does not acquire state from spans it never had reason to consult. Deployed practice routinely
assumes the converse --- that a downstream product may stand in for the upstream evidence it was
derived from. A note summarizing a retrieved document is treated as licence to evict the document;
a citation attached to an answer sentence is treated as the record of what produced it. We test
that assumption where it is load-bearing in two such heuristics, on one set of agent trajectories,
and it inverts in both.

\textbf{Setup.} We run $247$ tool-using trajectories over HotpotQA \citep{hotpotqa2018} with
Qwen3-8B, labelling seven regions of the resulting stream (system, user, plan, tool input, tool
output, scratchpad, answer). Readout is the log-odds of a gold value against a matched decoy under
forced choice; ablation drops a region's cache rows from the served set while every retained row
keeps its original position, so nothing is re-encoded. Follow-up questions come in two kinds:
those whose answer the agent's own notes already state, and those about a value present in the
retrieved text that the agent never wrote down. The second kind is the diagnostic one --- it asks
whether anything other than the source still carries the value.

\textbf{Prospectively, the measured priority table is the inverse of the hand-set one.} Dropping
the retrieved documents costs $21.99$ nats on unstated follow-ups; dropping any other region costs
approximately nothing (plan $+1.52$, scratchpad $+0.81$, tool input $+0.09$, answer $-2.24$).
Region-aware eviction schedules assign retrieved documents a low base priority and a fast decay
rate by default, on the reasoning that low-value retrievals and stale tool outputs are useful only
for a short span of the trajectory \citep{memdecay2026}; on this measurement the retrieved documents
are the only load-bearing region. The attention half-lives fitted alongside that schedule already
disagree with it --- retrieval is the longest-lived movable region in every setting measured, against
its own fast-decay default --- so the hand-set table is contradicted by two independent instruments
in the same direction. The agent's own
output does carry source values, but almost entirely in the cases where it wrote them down
($+1.77$ when the notes state the value, $+0.38$ when they do not), and the \emph{plan} --- written
before the search ran --- shows exactly zero dependence on the source value it precedes
($+0.000$, $p=1.0$). Causal order, not content, decides what a region can carry. The same boundary
governs what an agent's own answer can stand in for: after a turn, keeping the answer's KV and
dropping the retrieved pages recovers follow-ups about facts the answer stated at the full-context
rate ($1.00$, $n{=}216$) and follow-ups about retrieved facts it did not state at $.39$
($n{=}244$), against $.97$ for re-retrieval; keeping the model's reasoning tokens as well adds
nothing ($+0.53$ log-odds, $p{=}.32$). An answer's KV covers what it stated and no more.

\textbf{Why dropping the source rarely costs accuracy here.} Drop the tool output in which the answer first appears and $242$ of the $247$ trajectories still read out the gold value over its matched decoy; all $242$ do so as well in a counterfactual stream where that output had carried a different value and every later token is byte-identical. In $232$ of them the value survives as text elsewhere --- in the agent's notes, its answer, or another tool output. The cache channel itself is small: with the downstream text fixed, the original-minus-counterfactual difference in gold log-odds is $+1.35$ on average (median $+0.5$; placebo $-0.03$), against $+32.3$ when the dropped output alone is served. On real trajectories an unchanged accuracy therefore mostly reflects textual copies; the $99{:}0$ of \S\ref{sec:discovery} is an existence proof under controlled donor pairs, not an estimate of how much a natural trajectory leans on the cache.

\textbf{The attention signal available at eviction time marks only what need not be kept.} Within
the retrieved documents we rank $16$-token blocks and keep the top $B\%$, holding everything else
fixed. The heavy-hitter family scores blocks by accumulated attention
\citep{h2o2023,scissorhands2023,snapkv2024}; the question is whose attention. Attention from the
follow-up is not available when an agent must evict, because the follow-up has not arrived. What is
available is the attention the agent accumulated while generating its own plan, queries and notes.
Making the source of attention the only variable separates the two sharply
(Table~\ref{tab:consequences}). Attention accumulated during the agent's own generation is
significantly useful on stated follow-ups and indistinguishable from random on unstated ones;
attention from the follow-up shows no such asymmetry and, if anything, the opposite sign. The
mechanism is the contract read at token granularity: the agent attended to what it was using to
write, so those tokens are the ones whose content already left the cache in textual form --- and
are therefore the ones it need not retain. What must be retained is what it read and did not write,
and about that its attention is silent. The asymmetry survives stratifying items by where the
target sits inside the retrieved text, which a purely positional ranker does not.

Two readings follow. First, ``high accumulated attention marks what to keep'' is not a property of
attention but a conditional one: the same quantity, same model, same items and same budget yields
opposite verdicts depending on who attends. Second, the standard validation setting for eviction
--- single-turn long-context question answering --- places the question in the prompt, which makes
accumulated attention query-aware by construction. That is the column we reproduce. Eviction in a
multi-turn agent happens \emph{before} the next query, which is the other column.

\textbf{Retrospectively, the reported failure mode is the rarer one.} The same dependence test runs
backwards. Automatic citation evaluation scores whether a cited passage \emph{supports} a statement
\citep{alce2023,bohnet2022}; support is not use. \citet{wallat2024} separate the two and measure
post-rationalization, reporting up to $57\%$ unfaithful citations; the complementary condition they
define --- change the value in a cited passage and the statement should change --- is the one a
value swap runs directly. On $418$ cited multi-hop answers, swapping the value in the answer-bearing
passage moves the readout by $+36.1$ nats on average, against $+0.04$ for a placebo passage
($1\%$ of items exceed threshold) and $+40.7$ for a planted control the model cannot know
($100\%$). Under that probe, decorative citation --- cited but not load-bearing --- accounts for
$1.5\%$ of citations. The dominant failure is the opposite one: $20.7\%$ of load-bearing passages
were used and not cited, and the values still move by $+30.7$ nats on exactly those items. The
concentration is structural. Bridge passages, which supply the entity linking the two hops, are
load-bearing on $76\%$ of items and cited on $30\%$. An entailment check cannot ask for them,
because a bridge passage does not entail the answer sentence --- it is upstream of it. The
heuristic under test is again that the downstream product represents its upstream evidence.

\textbf{One quantity, two directions.} Eviction asks prospectively which rows may be dropped;
attribution asks retrospectively which rows earned credit. They are the same measurement run
forwards and backwards, and both defaults invert for the same reason. Neither result is an
algorithm: we compare ranking signals under one protocol rather than reimplementing eviction
systems, and we report no end-to-end accuracy under a budget. What transfers is the measurement and
its direction.

\textbf{Scope.} One checkpoint; trajectories are shallow (median two search steps, about four
retrieved passages); unstated probes are years, a highly extractive fact type, so the density
profile of composite facts is likely flatter. Our attention arms average over heads and layers into
one ranking, whereas H2O and SnapKV account and evict per head; H2O also pairs heavy hitters with a
recent window over the sequence \emph{tail}, which these trajectories never evict, so that component
is untested here. Two query-free lexical rankers score reliably \emph{below} random and we could not
determine why: the effect is stable across target position and across whether the ranker has any
lexical overlap with the probed passage.

\subsection{Related work}
\label{app:ext-related}

\textbf{The nearest mechanism work, and where ours departs.} Models Take Notes \citep{mtn2026}
already shows prefill writing memoized downstream conclusions, field-level note editing, an
append-only erratum, and position-portable notes that can be RoPE-repositioned and spliced
elsewhere; MEMENTO \citep{memento2026} is the closest \emph{source-omitted} analogue, training a
model to emit compressed ``mementos'' and finding that a memento's KV retains implicit information
from the masked reasoning block --- ablating that channel costs $15$ points. KVEraser
\citep{kveraser2026} shows the same residue persisting in suffix KV after deletion and learns to
steer it away; the lookback analysis of \citet{lookbacks2025} traces the pointer-style
address/payload mechanism by which models track state in context. We do not re-discover downstream
notes. What we isolate is \emph{source-omitted sparse event serving} --- what an independently
served span carries once its source is dropped, with the future query absent at write time and a
frozen model reading through its native interface --- reached from the opposite direction to
MEMENTO: unmodified checkpoints, answer-free spans rather than trained summaries, and donor-pair
identification with served tokens and positions held fixed. Two contemporaneous preprints take
other operating points on trajectory KV: KV-PRM \citep{kvprm2026} transfers the full generated
cache to a task-specific verifier, and AAFLOW+ \citep{aaflow2026} treats KV materialization and
transfer as distributed runtime operators. The senses are layered --- physical KV objects,
task-specific readers over full caches, and write-time semantic commitment into independently
servable spans (this work).

\textbf{KV reuse and repair: that line repairs the rows it keeps; we audit what the omitted source
left behind.} Four lines of work manage, repair, or compose reused KV. \emph{Cross-chunk
recomputation} --- CacheBlend \citep{cacheblend2024}, EPIC \citep{epic2024}, InfoFlow~KV
\citep{infoflowkv2026} --- concatenates independently encoded chunks and selectively recomputes the
positions carrying cross-chunk dependency, and KEEP \citep{keep2026} adds memory grouping,
multi-hop reconstruction of cross-attention between groups, and layer-balanced KV loading.
\emph{Learned link tokens} (KVLink, \citealp{kvlink2025}) restore self-attention across separately
cached chunks; \emph{eviction/compression} (SnapKV, \citealp{snapkv2024}; H2O,
\citealp{h2o2023}) shrinks the retained row set. HYPIC \citep{hypic2026} and C$^2$KV
\citep{c2kv2026} extend position-independent reuse to hybrid-attention models and to compressed,
composable non-prefix segments. All optimize the \emph{physical} reuse of rows they keep; whether
such composed states preserve source-omitted \emph{semantic} materialization remains unmeasured.
Our source-present multi-hop denoising ceiling (App.~\ref{app:tables}) is exactly the regime this
line optimizes.

\textbf{Memory systems build the shell; we measure the substrate.} EpiCache \citep{epicache2025}
selects and compresses episodic KV under query-time uncertainty on the same REALTALK/LoCoMo
distributions, but \emph{retains} selected original-history episodes for later selection; we
\emph{omit} the source and audit what answer-free downstream spans carry. Zep \citep{zep2025} and
APEX-MEM \citep{apexmem2026} provide temporal knowledge-graph and semi-structured agent memory,
MemGPT \citep{memgpt2023} and EM-LLM \citep{emllm2024} manage episodic context, and long-context
stress tests \citep{ruler2024} supply further serving distributions. Our contribution to this line
is not another shell but the substrate-level contract those shells would need if they served
event KV.

\subsection{Limitations and scope}
\label{app:ext-limits}

We demonstrate a programmable, per-checkpoint sparse-memory primitive, not a universal latent write
API. The donor swaps certify a causal channel for \emph{compact} state only: verbatim payloads
rarely transfer ($9.7\%$), exact numeric payloads are not natively recovered, derived conclusions
need a write-time prompt that itself costs full-context accuracy ($.818\to.698$), and the flat
multi-hop curve uses oracle row selection, making it a mechanism ceiling rather than a deployed
number. \S\ref{sec:address} measures what that assumption was holding up instead of leaving it as a
hedge, and the resulting negative is deliberately bounded: on the synthetic template it covers
bag-of-words and a dense bi-encoder; on real dialogs it covers the four text stages a deployed
stack runs --- BM25, a dense bi-encoder, rank fusion and a cross-encoder reranker --- at rank
level, on one corpus of ten conversation clusters. A directed graph index was run but failed its
instrument gate, and no multi-hop router was tested, so whether a structured retriever can address
materialized state remains open rather than settled. The synthetic retrieval rates are properties
of the template, not forecasts: what generalises is that an index over event \emph{text}
cannot represent the direction of a reference edge, and therefore cannot separate the row that
carries derived state from the row that points at it.
Every negative accessibility verdict on the synthetic envelope is an output-interface measurement,
and representation-level probes now adjudicate it (App.~\ref{app:repprobe}): they find the state in the served rows where the readout fails, so those verdicts bound reading rather than storage, though decodability is not use; the real-dialog null has its own probe
(\S\ref{sec:active}), and it reads diffuse, content-blind drift. On real dialogs the null was first
scoped to our carrier-selection heuristic and recognition readout; it now also covers rows selected
by a deployed retrieval stack, a log-odds readout, chain-of-thought and the deployed hot window ---
all on Qwen3-8B and LoCoMo's ten conversation clusters (REALTALK's ten dyads form three participant
components, so LoCoMo carries the inference), with wild-cluster bootstrap-$t$ inference because at
ten clusters the percentile cluster bootstrap over-rejects. Withholding the value-bearing turns is a
construction on single-hop questions, where the stack finds them, and the natural condition on
multi-hop ones (\S\ref{sec:address}). Three checks interrogate the serving path itself: compact assembly
reproduces one-shot prefill on full-attention Qwen3 (App.~\ref{app:protocols}); on Gemma-4's
sliding-window layers it does not, so that arm is served at original positions instead
(App.~\ref{app:swgate}); and $16$ of $20$
conversations exceed Qwen3's native window, where restoring range with YaRN \emph{widens} the null
rather than closing it (both arms re-run together on one stack: LoCoMo $-.048$ $[-.103,+.009]$
without scaling, $-.121$ $[-.181,-.068]$ with it; paired per-conversation change $-.074$,
$p{=}.02$). Finally, released
linear-attention hybrids did not, as we loaded them, expose token-addressable KV at every layer
\citep{qwen36_2026}, so row-serving does not apply unchanged --- a scope limit that strengthens the
case for explicit materialization events \citep{hypic2026}.

\end{document}